\documentclass{fairmeta}

\usepackage{amsmath}
\usepackage{amssymb}
\usepackage{amsfonts}
\usepackage{algorithm}
\usepackage{algpseudocode}
\usepackage{makecell}
\usepackage{wrapfig}
\usepackage{bbm}
\usepackage{enumitem}
\usepackage{nicefrac}
\usepackage{pifont}
\usepackage{colortbl}
\usepackage{array}

\graphicspath{{figures/}}

\definecolor{topone}{RGB}{223,240,216}
\definecolor{toptwo}{RGB}{220,226,246}
\definecolor{topthree}{RGB}{245,238,204}

\usepackage{amsmath}
\usepackage{xcolor}
\usepackage{booktabs,multirow,array}

  \DeclareMathAlphabet\mathbfcal{OMS}{cmsy}{b}{n}

  \def\0{{\bf 0}}
  \def\1{{\bf 1}}

\def\ie{\emph{i.e.}}

\title{OccamToken: Efficient VLM Inference with Training-Free and Budget-Adaptive Token Pruning}
\author[1,*]{Geng Li}
\author[1,*]{Guohao Chen}
\author[1,*]{Ting Chen}
\author[1]{Shilin Shan}
\author[1]{Kuangji Zuo}
\author[1]{Bofan Lyu}
\author[1]{Tuo An}
\author[1,\dagger]{Gen Li}
\author[1,\dagger]{Jianfei Yang}
\affiliation[1]{Nanyang Technological University (NTU)}
\contribution[*]{Equal contribution.}
\contribution[\dagger]{Corresponding authors.}

\abstract{
Vision-language models (VLMs) rely on long visual token sequences for visual understanding, making the prefill stage expensive in both computation and memory.
Most existing pruning methods follow an absolute-ranking paradigm, assigning importance scores to visual tokens and retaining a fixed top-$K$ subset.
In this work, we argue that this paradigm is fundamentally brittle: attention sinks distort token importance rankings, while image redundancy and query-dependent visual evidence make fixed token budgets unreliable across inputs.
We propose \textbf{OccamToken}, a training-free framework that replaces absolute token ranking with \textit{register-anchored relative evidence testing}.
Instead of asking which tokens are globally important, OccamToken evaluates whether a visual token provides information beyond a register-based reference.
Our key insight is that register tokens naturally absorb low-information attention patterns, making them a stable reference for identifying genuinely informative visual evidence.
Based on this principle, OccamToken performs both image-adaptive redundancy pruning and query-adaptive relevance pruning through dynamic thresholds derived from register attention.
Across LLaVA-NeXT, LLaVA-v1.5, and Qwen3-VL, OccamToken consistently improves the accuracy--efficiency trade-off without additional training.
Notably, on LLaVA-NeXT, it reduces 2,880 visual tokens to approximately 40 while preserving over 93\% of the original accuracy, enabling stable visual token compression even in the extreme 1.4\% retention regime.
}

\begin{document}

\maketitle
\section{Introduction}

Vision-language models (VLMs)~\citep{liu2024improved,bai2023qwen,li2025mini,liu2024llavanext} have achieved remarkable progress in multimodal understanding and reasoning, as evidenced by recent general-purpose VLMs and multimodal evaluation benchmarks~\citep{liu2024mmbench,yue2024mmmu,li2023evaluating}.
A standard VLM connects a vision encoder with a large language model by encoding images into patch-level visual tokens, projecting them into the language embedding space, and prepending them to textual instructions~\citep{liu2023visual,dai2023instructblip}.
As high-resolution, multi-image, and video-capable VLMs become increasingly common, each input may contain hundreds or even thousands of visual tokens~\citep{liu2024llavanext,bai2025qwen3}.
These tokens are processed during the prefill stage and stored in the key-value cache for autoregressive decoding, making visual sequence length a major source of computation and memory overhead~\citep{vaswani2017attention,dao2022flashattention,chen2024image}.
This motivates visual token compression: reducing redundant visual tokens while preserving the evidence required for multimodal reasoning~\citep{shang2025llava,zhang2025sparsevlm,yang2025visionzip,zhangbeyond}.

Existing compression methods typically follow a \emph{score-and-select} paradigm: they assign importance or redundancy scores to visual tokens and retain a compact subset, often under a fixed top-$K$~\citep{chen2024image} budget.
The scoring signals can be broadly categorized into attention-based cues, such as \texttt{[CLS]}-to-patch or \texttt{[text]}-to-vision attention~\citep{chen2024image,zhang2025sparsevlm,zhang2025beyond,yang2025visionzip}, and feature-level cues, such as token similarity, diversity, duplication, or set coverage~\citep{alvar2025divprune,wen2025stop,dengscope}.
Among these directions, attention-based scoring remains a dominant and widely adopted choice because it directly reuses the interaction patterns already computed by the vision encoder or the language model.
However, we observe that this attention-based pruning paradigm suffers from two structural deficiencies.

\begin{figure}[t]
\centering
\includegraphics[width=1\textwidth]{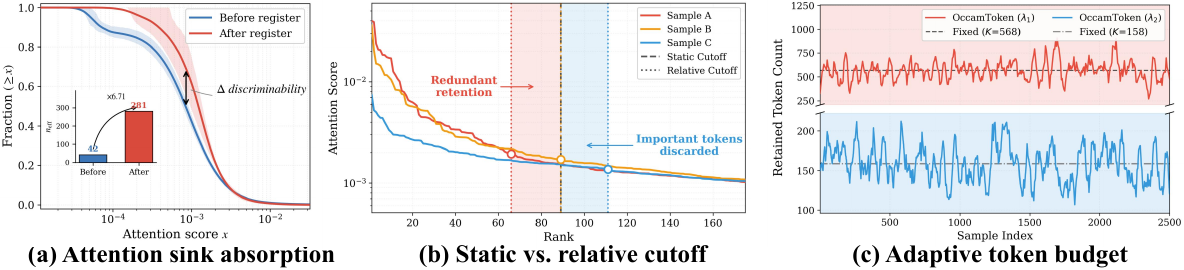}
\caption{\textbf{Motivation for relative comparison.}
\textbf{(a)}~Register insertion suppresses attention sinks, yielding a less sink-dominated CLS$\to$vision attention distribution and increasing $n_{\mathrm{eff}}$ from 42 to 281.
\textbf{(b)}~A fixed top-$K$ cutoff corresponds to different evidence levels across samples, while relative cutoffs adapt to distributional variation.
\textbf{(c)}~OccamToken produces sample-adaptive token budgets through two-stage register-anchored pruning.}
\label{analy}

\end{figure}

\textit{\textbf{First, low-information tokens can absorb excessive score mass and make visual-token importance less distinguishable.}}
For visual token pruning, the score distribution should ideally separate informative visual evidence from redundant or background regions.
However, a few visually uninformative tokens can receive unusually large attention scores, causing the remaining visual tokens to receive smaller and less distinguishable scores.
Prior work~\citep{barbero2025llms,yu2024unveiling} describes this phenomenon as \emph{attention sinks}, \ie, high-norm outlier tokens attract excessive attention.
Consequently, truly informative tokens are squeezed into a narrower score range, reducing score gaps and obscuring the boundary between informative and redundant tokens.
\textit{\textbf{Second, the static top-$K$ cutoff cannot adapt to the per-sample variation in importance-score distributions.}} 
In practice, attention-score distributions vary substantially across images and queries. 
A fixed $K$ therefore cannot adapt its retention budget to each sample, often preserving redundant tokens in simple cases while discarding critical evidence in complex ones. 
Existing adaptive methods either rely on extra training~\citep{ye2025atp,huangdynamic} or use training-free image-level statistics~\citep{fangprune,shang2025llava}, which remain weakly responsive to query-conditioned evidence demands.

We find that these two deficiencies can be jointly addressed by using register tokens as reference anchors.
Register tokens, introduced in recent transformer studies~\citep{darcet2024vision,jiangvision}, can aggregate sink-like attention scattered across low-information image regions.
Meanwhile, prior work shows that register tokens encode image-level global information~\citep{jiangvision}, and our analysis further suggests that they are less sensitive to fine-grained local patch evidence; see Section~\ref{sec:register_threshold}.
Thus, a register token is not merely a sink absorber, but also provides a global yet non-discriminative summary of the visual input.

This motivates a \emph{reference-adaptive pruning paradigm}: instead of asking whether a visual token ranks within a fixed top-$K$ set, we ask whether it provides stronger evidence than the register token under the current attention distribution. 
Because the register token is scored in the same attention competition as ordinary visual tokens, it serves as a semantic reference that both reduces sink-induced score distortion and defines a sample-adaptive pruning boundary. 
Building on this principle, we propose \textbf{OccamToken}, a training-free two-stage framework for reference-adaptive visual token pruning during VLM inference.
\textit{Stage~I} operates at the vision encoder output and removes image-level redundancy with \texttt{[CLS]}-to-register attention.
\textit{Stage~II} operates inside the language model and further prunes query-irrelevant visual tokens with \texttt{[text]}-to-register attention.
Our main contributions:

\begin{itemize}[leftmargin=*]
    \item We propose \textbf{register-anchored dynamic thresholding}, a training-free mechanism that links attention-sink mitigation to adaptive visual token pruning.
By inserting a test-time register token into the same attention competition, the mechanism absorbs sink-like attention from low-information tokens and uses the register score as a distribution-adaptive pruning threshold.

    \item We instantiate this mechanism as \textbf{OccamToken}, a training-free two-stage pruning framework for efficient VLM inference.
It first removes image-level redundancy at the vision-encoder output with \texttt{[CLS]}-to-register attention, and then prunes query-irrelevant tokens inside the language model with \texttt{[text]}-to-register attention, yielding image- and query-adaptive token budgets.

   \item Extensive experiments on image-understanding benchmarks across LLaVA-v1.5~\citep{liu2024improved}, LLaVA-NeXT~\citep{liu2024llavanext}, and Qwen3-VL~\citep{bai2025qwen3} show that OccamToken outperforms the compared state-of-the-art visual token pruning baselines, including both training-free and training-based methods, under matched token budgets.
Notably, OccamToken pushes training-free visual token compression on LLaVA-NeXT into the 1.4\% retention regime while preserving over 93\% of the full-token accuracy.
\end{itemize}

\section{Related Work}

\subsection{Visual Token Compression in VLMs}

The redundancy of visual tokens during VLM inference has motivated various compression strategies, including pruning, merging, and hybrid methods.
Existing approaches typically estimate token importance from attention signals~\cite{zhang2025beyond,shang2025llava,yang2025visionzip,chen2024image,ye2025fit,zhang2025sparsevlm}, or from feature-level criteria such as diversity, duplication, and coverage~\cite{alvar2025divprune,wen2025stop,dengscope,dhouib2025pact}.
Most of these methods still rely on fixed retained-token budgets, which cannot fully adapt to sample-dependent redundancy.

Recent adaptive methods attempt to relax this constraint by learning token-budget predictors or deriving budgets from training-free statistics.
However, learned approaches require additional parameters, data, or model-specific adaptation~\cite{ye2025atp,huangdynamic,takezoelearnpruner,shao2025growing}, while training-free statistical methods depend on externally defined criteria such as sparsity, similarity, or global redundancy scores~\cite{shang2025llava,fangprune}.
In contrast, OccamToken introduces a reference-based adaptive mechanism: the register token participates in the same softmax competition as visual tokens, making its score an internal semantic reference for dynamic pruning.
This enables training-free, query-adaptive token allocation without learned predictors or externally defined statistical cutoffs.

\subsection{Attention Sinks and Register Tokens}

Attention sinks are a common Transformer phenomenon in which a small number of tokens attract disproportionately large attention weights~~\citep{sunmassive,guattention,cancedda2024spectral}.
Xiao et al.~\cite{xiaoefficient} studied this behavior in language models, while Darcet et al.~\cite{darcet2024vision} observed high-norm outlier tokens in low-information regions of vision Transformers and introduced trained register tokens to absorb such artifacts.
Jiang et al.~\cite{jiangvision} further showed that similar behavior can be induced at test time by redirecting sink-like activations to an appended register token, avoiding retraining.

Prior work primarily uses register tokens to stabilize Transformer representations or mitigate attention-sink artifacts.
OccamToken instead repurposes the register token as a semantic reference anchor for visual token pruning.
By using the register score as an adaptive threshold, our method connects attention-sink mitigation with dynamic token budgeting in a training-free and plug-and-play manner.

\section{Method}

\subsection{Preliminaries}
\label{sec:preliminaries}

\textbf{VLM inference pipeline.}
Given an image $x$ and a text query, a vision encoder~\citep{radford2021learning,zhai2023sigmoid}~$g$  first maps the image into a sequence of visual tokens
$\mathcal{V}=g(x)=\{v_i\}_{i=1}^{N} \in \mathbb{R}^{N \times d_v}$,
together with a \texttt{[CLS]} token $v_{\mathrm{cls}}$.
A visual projector~\citep{dai2023instructblip}~$p$ then maps these visual representations into the language-model embedding space.
Let $\mathcal{T}=\{t_j\}_{j=1}^{M}\in\mathbb{R}^{M\times d}$ denote the embedded text tokens.
The language model takes the concatenated sequence
$[p(\mathcal{V});\,\mathcal{T}]$
as input and performs autoregressive decoding.
Since the number of visual tokens is often much larger than the number of text tokens, \ie, $N \gg M$, visual tokens dominate the prefill-stage computation and memory cost, making the compression of $\mathcal{V}$ critical for efficient VLM inference.

\textbf{Attention mechanism.}
Both the vision encoder and the language model are built on Transformer blocks, whose core operation is multi-head attention.
For clarity, we describe a single attention head.
Given an input sequence $\mathbf{X}\in\mathbb{R}^{L\times d}$, the attention weights and output are computed as
\begin{equation}
    \mathbf{A}
    =
    \mathrm{Softmax}\!\left(
    \frac{\mathbf{Q}\mathbf{K}^{\top}}{\sqrt{d_h}}
    \right),
    \qquad
    \mathbf{O}=\mathbf{A}\mathbf{U},
\end{equation}
where
$\mathbf{Q}=\mathbf{X}\mathbf{W}_q$,
$\mathbf{K}=\mathbf{X}\mathbf{W}_k$,
and
$\mathbf{U}=\mathbf{X}\mathbf{W}_u$
denote the query, key, and value projections, respectively, and $d_h$ is the head dimension.
The entry $A_{ij}$ measures how strongly token $i$ attends to token $j$.
Existing visual token pruning methods often use attention weights from specific evaluator tokens, such as \texttt{[CLS]} or \texttt{[text]} tokens, to visual tokens as importance scores.

\textbf{Test-time register token.}
Jiang et al.~\cite{jiangvision} trace attention sinks to sparse \textit{register neurons} in the MLPs and introduce a test-time register token $r$ to absorb their high-norm activations.
During inference, abnormal channel-wise activations from patch tokens are redirected to the corresponding dimensions of $r$, and $r$ then participates in subsequent self-attention computations.
Thus, the register token serves as a sink recipient while being scored in the same softmax distribution as ordinary visual tokens.

\begin{figure}[t]
\centering
  \includegraphics[width=1\textwidth]{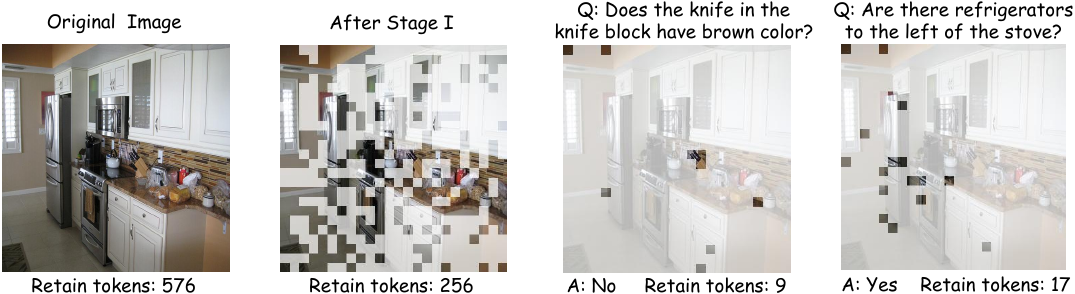}
\caption{
\textbf{Visualization of query-adaptive token budgets.}
For the same image, Stage~I performs query-agnostic redundancy pruning, while Stage~II selects query-relevant visual evidence.
Different questions induce different retained token budgets: the knife-block question keeps 9 tokens, whereas the refrigerator-stove relation question keeps 17 tokens.
This demonstrates that OccamToken adapts the final visual budget to query-specific evidence demand rather than enforcing a fixed top-$K$ budget.
}
  \label{fig:query_adaptive_main}

\end{figure}

\subsection{From Absolute Ranking to Relative Comparison}
\label{sec:motivation}

The goal of visual token pruning is to select a compact subset 
$\mathcal{S} \subset \mathcal{V}$ such that the model output conditioned on 
$\mathcal{S}$ closely approximates that conditioned on the full visual token set 
$\mathcal{V}$.
Given an importance scoring function $s$, a widely adopted pruning rule is to rank visual tokens by their scores and retain the top-$K$ tokens:
\begin{equation}
    \mathcal{S} = \mathrm{top}\text{-}K(\mathcal{V}, s).
\end{equation}
While simple and effective, this top-$K$ ranking rule exposes two limitations.

\textbf{Observation 1: Attention sinks reduce the separability of importance scores.}
Due to the sum-to-one constraint of softmax attention, attention allocation is inherently competitive.
Attention sink tokens can absorb a disproportionate fraction of this fixed attention mass, leaving the remaining visual tokens with compressed scores and smaller score gaps.
This makes informative tokens less distinguishable from redundant ones.
We quantify this effect using the effective number of attended tokens,
\ie, $n_{\mathrm{eff}} {=} \exp(H(\boldsymbol{\alpha}))$, where 
$H(\boldsymbol{\alpha}) {=} -\sum_i \alpha_i \log \alpha_i$ is the entropy of the attention distribution.
Here, $n_{\mathrm{eff}}$ corresponds to the support size of a uniform distribution with the same entropy.
As shown in Figure~\ref{analy}~(a), before introducing a register token to absorb sink-like activations, $n_{\mathrm{eff}}$ is only $42$.
After sink absorption, it increases to $281$, indicating a substantially less sink-dominated scoring distribution.

\textbf{Observation 2: A fixed top-$K$ cutoff is mismatched with varying score distributions.}
Top-$K$ pruning imposes a rank-based decision rule: the token at the $K$-th rank defines the pruning boundary.
However, even after sink artifacts are mitigated, the attention-score distribution can still vary substantially across images and queries, as shown in Figure~\ref{analy}~(b).
As a result, the same value of $K$ may correspond to very different evidence levels across samples.
For information-dispersed samples, it may discard critical evidence; for information-concentrated samples, it may retain unnecessary redundancy.
We refer to this issue as a \textit{cutoff-distribution mismatch}: a static rank-based cutoff cannot adapt to sample-dependent score distributions.

These observations motivate moving beyond absolute ranking.
Instead of asking whether a token ranks among the top $K$, pruning should ask whether its score exceeds a meaningful reference level under the current score distribution.
We therefore propose a \textit{relative comparison paradigm}, which replaces the static cutoff with an adaptive reference score produced within the attention competition:
\begin{equation}
\label{eq:framework}
    \mathcal{S}(x)
    =
    \left\{
    v \in \mathcal{V}(x)
    \mid
    s_x(v) \geq \lambda \cdot s_x(r(x))
    \right\},
\end{equation}
where $x$ denotes the input sample, $r(x)$ is a sample-dependent reference token, and $\lambda$ controls the pruning intensity.
Under this criterion, a visual token is retained not because of its global rank, but because it provides stronger evidence than the reference token.
The number of retained tokens $|\mathcal{S}(x)|$ therefore varies with the score distribution of each sample.

 \begin{figure}[t]
  \centering
  \includegraphics[width=1\textwidth]{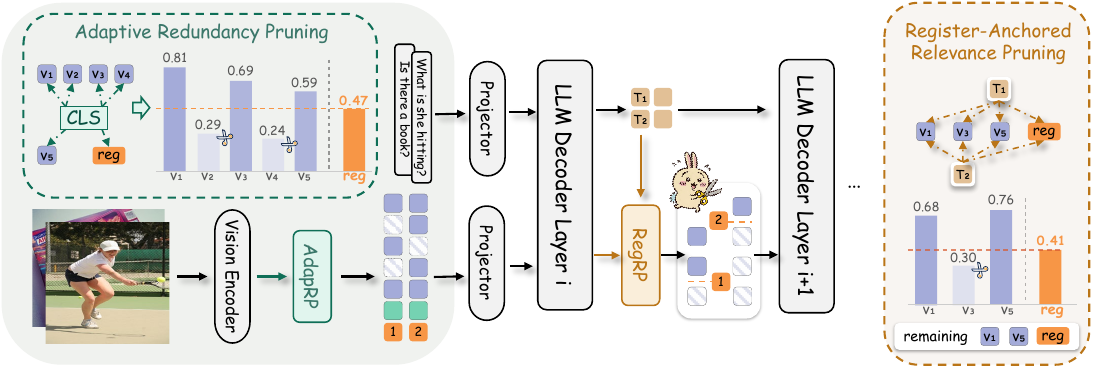}
  \caption{\textbf{Overview of OccamToken.} Given an input image and a text query, our framework performs two-stage adaptive visual token pruning. Stage~I \textbf{AdapRP} (\textbf{Adap}tive \textbf{R}edundancy \textbf{P}runing): At the vision encoder output, a test-time register token absorbs attention sinks. The \texttt{[CLS]} token scores all visual tokens and the register token jointly; tokens scoring below $\lambda_1 \cdot s_1(r)$ are pruned, yielding an image-adaptive token budget. Stage~II \textbf{RegRP} (\textbf{Reg}ister-\textbf{A}nchored \textbf{R}elevance \textbf{P}runing): Within the LLM, text tokens score the surviving visual tokens via max attention, while the register token's mean text attention serves as a dynamic threshold. Tokens scoring below $\lambda_2 \cdot s_2(r)$ are removed, producing a query-adaptive final budget.}
\label{fig:occamtoken_overview}

\end{figure}

\subsection{Register Token as a Dynamic Reference Anchor}
\label{sec:register_threshold}

Eq.~\ref{eq:framework} suggests that adaptive pruning requires a reference token whose score can serve as a sample-dependent threshold.
However, selecting such a reference is non-trivial.
An ideal reference should satisfy three properties:
\textit{(i)} it should be introduced without additional training;
\textit{(ii)} it should participate in the same softmax attention computation as visual tokens, so that its score is directly comparable to visual-token scores;
and \textit{(iii)} it should represent non-discriminative visual information, providing a meaningful cutoff between informative and uninformative tokens.

The first two properties are in fact implicit in Observation~1.
The register token redirects high-norm sink activations at inference time, increasing $n_{\mathrm{eff}}$ from $42$ to $281$ without model retraining.
Moreover, once included in the token sequence, it participates in the same softmax attention computation as ordinary visual tokens, yielding a sample-specific score directly comparable to visual-token scores.

\begin{center}
    \textit{Can the register token provide a meaningful non-discriminative reference?}
\end{center}

We answer this question by examining what information the register token encodes.
First, prior work shows that register tokens capture image-level global information.
Jiang et al.~\cite{jiangvision} conduct linear probing experiments and find that the register token achieves classification accuracy comparable to the \texttt{[CLS]} token, suggesting that it encodes global visual semantics.

Second, we observe that the register token contains little fine-grained local patch evidence.
To verify this, we rank image patches by their \texttt{[CLS]}-to-patch attention scores on LLaVA-v1.5~\citep{liu2024improved}, then mask either the top $50\%$ or bottom $50\%$ of patches.
We compute the cosine similarity of the register representation before and after masking, obtaining $0.981$ and $0.967$, respectively.
The register representation remains largely stable regardless of which half of the patches is removed, indicating that it is insensitive to specific local patch evidence.

Together, these findings suggest that the register token encodes a global summary rather than localized discriminative content.
This gives it a useful semantic role: it acts as a global but non-discriminative reference anchor.
Comparing a visual token with the register token therefore asks whether the local patch provides stronger evidence than this global reference.
Accordingly, Eq.~\ref{eq:framework} retains tokens whose scores exceed the register score, enabling sample-adaptive pruning without a fixed top-$K$ cutoff.

\subsection{Stage~I: Image-Adaptive Redundancy Pruning}
\label{sec:stage1}

Following the \textit{relative comparison paradigm} introduced in Section~\ref{sec:register_threshold}, we evaluate each visual token relative to the register token rather than ranking tokens by an absolute top-$K$ cutoff.
We instantiate this principle in two stages.
\textbf{Stage~I} operates at the output of the vision encoder, where the \texttt{[CLS]} token serves as the evaluator to remove redundant visual tokens in an image-adaptive manner.
\textbf{Stage~II} operates inside the language model, where text tokens serve as query-conditioned evaluators to further prune visual tokens that are irrelevant to the current query.

\textbf{\texttt{[CLS]} scoring and dynamic pruning.}
After constructing the register token as described in Section~\ref{sec:register_threshold}, we use the \texttt{[CLS]} token to jointly score all visual tokens and the register token:
\begin{equation}
\label{EQ:CLS}
    s_1(v_i) = A_{\mathrm{cls} \to v_i},
    \qquad
    s_1(r) = A_{\mathrm{cls} \to r},
    \qquad
    \forall v_i \in \mathcal{V}.
\end{equation}
The register score then serves as an image-adaptive reference threshold.
Specifically, the retained token set after Stage~I is defined as
\begin{equation}
\label{eq:stage1}
    \mathcal{S}_1
    =
    \{v_i \in \mathcal{V} \mid s_1(v_i) \geq \lambda_1 \cdot s_1(r)\}
    \cup \{r\},
\end{equation}
where $\lambda_1$ controls the pruning intensity, and the register token is always retained for Stage~II.
Since the threshold is derived from the current image's attention distribution, the number of retained tokens $|\mathcal{S}_1|$ varies with image content, enabling image-adaptive budget allocation.

\textbf{Generalization to diverse architectures.}
Some VLMs~\citep{bai2025qwen3,bai2023qwen} do not provide a \texttt{[CLS]} token in the vision encoder.
To support such architectures, we introduce a \textbf{\textit{mutual-scoring scheme}} that uses visual tokens themselves as evaluators.
All visual tokens and the register token attend to each other, and the importance of each candidate token is measured by the average attention it receives:
\begin{equation}
\label{eq:mutual_scoring}
\begin{aligned}
    s_1^{\mathrm{mutual}}(v_i)
    &=
    \frac{1}{|\mathcal{V}|+1}
    \sum_{u \in \mathcal{V} \cup \{r\}}
    A_{u \to v_i},
    \\
    s_1^{\mathrm{mutual}}(r)
    &=
    \frac{1}{|\mathcal{V}|+1}
    \sum_{u \in \mathcal{V} \cup \{r\}}
    A_{u \to r}.
\end{aligned}
\end{equation}
The Stage~I pruning rule is then applied analogously, using
$\lambda_1 \cdot s_1^{\mathrm{mutual}}(r)$ as the dynamic threshold.
We validate this scheme on Qwen~\citep{wang2024qwen2,bai2025qwen3}, showing that register-anchored thresholding remains effective even for vision encoder architectures without a \texttt{[CLS]} token; see Section~\ref{sec:experiments}.

\begin{algorithm}[t]
\caption{OccamToken: Two-Stage Adaptive Visual Token Pruning}
\label{alg:occamtoken}
\small
\begin{algorithmic}[1]
\setstretch{1.15}
\Require Image $I$, text query $T$, coefficients $\lambda_1,\lambda_2$, pruning layer $l$, register neuron set $\mathcal{K}_{\mathrm{reg}}$
\Ensure Model response with adaptively pruned visual tokens

\State Encode $I$ into visual tokens $\mathcal{V}$ and obtain MLP activations $h$
\State For $k\in\mathcal{K}_{\mathrm{reg}}$: 
$h_{r,k}\leftarrow \max_{v_i\in\mathcal{V}} h_{i,k}$, 
$h_{i,k}\leftarrow 0,\ \forall v_i\in\mathcal{V}$
    \Comment{redirect sinks to $r$}

\Statex \textbf{Stage~I: Image-Adaptive Redundancy Pruning}
\State Compute $s_1(v_i),s_1(r)$ via \texttt{[CLS]} attention; 
$\mathcal{S}_1 \leftarrow \{v_i \in \mathcal{V} \mid s_1(v_i) \geq \lambda_1 s_1(r)\} \cup \{r\}$
    \Comment{Eq.~\eqref{eq:stage1}}

\Statex \textbf{Stage~II: Register-Anchored Relevance Pruning}
\State Run LLM to layer $l$ with $[\mathcal{S}_1;\,T]$; compute $s_2(v_i),s_2(r)$ from text attention
    \Comment{Eq.~\eqref{eq:text_scoring}}
\State $\mathcal{S}_2 \leftarrow \{v_i \in \mathcal{S}_1 \setminus \{r\} \mid s_2(v_i) \geq \lambda_2 s_2(r)\} \cup \{r\}$
    \Comment{Eq.~\eqref{eq:stage2}}

\State \Return Continue autoregressive decoding with $[\mathcal{S}_2;\,T]$
\end{algorithmic}
\end{algorithm}

\subsection{Stage~II: Register-Anchored Relevance Pruning}
\label{sec:stage2}

After Stage~I, the retained set $\mathcal{S}_1$ has been filtered for image-level redundancy, but not all surviving tokens are necessarily relevant to the current query.
Stage~II therefore operates inside the language model, where cross-modal attention is used to further remove query-irrelevant visual tokens.

\textbf{\texttt{[Text]}-to-vision scoring.}
The retained set $\mathcal{S}_1$ is concatenated with the text tokens $\mathcal{T}$ and fed into the language model.
At the $l$-th decoder layer, we extract the attention scores from text tokens to tokens in $\mathcal{S}_1$.
For the register token, which provides a global but non-discriminative reference, we use the \textbf{\textit{mean}} text-to-register attention to obtain a stable reference score.
For each visual token, we use the \textbf{\textit{maximum}} attention it receives from text tokens, since a visual token may be relevant to only a few query words and averaging could dilute such localized correlations:
\begin{equation}
\label{eq:text_scoring}
\begin{aligned}
    s_2(r)
    &=
    \frac{1}{M}\sum_{j=1}^{M} A^{(l)}_{t_j \to r},
    \\
    s_2(v_i)
    &=
    \max_{j=1,\dots,M} A^{(l)}_{t_j \to v_i},
    \quad
    \forall\, v_i \in \mathcal{S}_1 \setminus \{r\}.
\end{aligned}
\end{equation}

Using $\lambda_2 \cdot s_2(r)$ as the query-adaptive threshold, the final retained set is defined as
\begin{equation}
\label{eq:stage2}
    \mathcal{S}_2
    =
    \{v_i \in \mathcal{S}_1 \setminus \{r\}
    \mid
    s_2(v_i) \geq \lambda_2 \cdot s_2(r)\}
    \cup \{r\},
\end{equation}
where $\lambda_2$ controls the pruning intensity.
Because $s_2(r)$ is computed from the current query-conditioned attention distribution, the threshold varies across queries.
As a result, the same image can yield different final token budgets for different questions, enabling query-adaptive budget allocation.

\begin{table}[!t]
\centering
\captionsetup{font=small, width=0.98\textwidth}
\resizebox{0.99\textwidth}{!}{
\begin{tabular}{llcccccccc|c}
\toprule
\textbf{Method} & \textbf{Type} & \textbf{GQA} & \textbf{SQA$^{\text{I}}$} & \textbf{VQA$^{\text{T}}$} & \textbf{POPE} & \textbf{MME} & \textbf{VQA$^{\text{v2}}$} & \textbf{MMB} & \textbf{MMB$^{\text{CN}}$} & \textbf{RelAcc.} \\
\midrule
\rowcolor{gray!20} \multicolumn{11}{c}{\textit{Upper Bound, 576 Tokens (100\%)}} \\
Vanilla & - & 61.9 & 69.5 & 58.2 & 85.9 & 1862 & 78.5 & 64.7 & 58.3 & 100.0\% \\
\midrule

\rowcolor{gray!20} \multicolumn{11}{c}{\textit{Retain Averaged 128 Tokens ($\downarrow$ 77.8\%)}} \\
\midrule
VisionZip $\ddagger$ & \textsc{Learned} & 58.9 & 68.3 & 57.0 & 83.7 & \underline{1823} & 76.6 & 62.6 & - & 97.3\% \\
TwigVLM & \textsc{Learned} & \underline{60.6} & \textbf{69.5} & \underline{57.8} & \underline{86.6} & 1818 & \textbf{77.9} & 63.5 & - & \underline{99.0\%} \\
LearnPruner & \textsc{Learned} & 60.3 & 68.5 & 57.3 & \textbf{86.7} & 1820 & \underline{77.3} & \underline{63.8} & 56.8 & 98.5\% \\
FastV & \textsc{Fixed} & 49.6 & 60.2 & 50.6 & 59.6 & 1490 & 61.8 & 56.1 & 51.4 & 82.1\% \\
SparseVLM & \textsc{Fixed} & 56.0 & 67.1 & 54.9 & 80.5 & 1696 & 73.8 & 60.0 & 51.1 & 92.6\% \\
DivPrune & \textsc{Fixed} & 59.3 & 69.0 & 56.1 & \textbf{86.7} & 1718 & 76.0 & 62.0 & 54.8 & 96.4\% \\
DART & \textsc{Fixed} & 57.9 & \underline{69.1} & 56.3 & 80.4 & 1721 & 74.7 & 60.7 & \underline{57.3} & 95.4\% \\
VisPruner & \textsc{Fixed} & 58.2 & \underline{69.1} & 57.0 & 84.6 & 1794 & 75.8 & 62.7 & \underline{57.3} & 97.2\% \\
VisionZip & \textsc{Fixed} & 57.6 & 68.9 & 56.8 & 83.2 & 1762 & 75.6 & 62.0 & 56.7 & 96.3\% \\
PruMerge+ & \textsc{Img.} & 58.2 & \underline{69.1} & 54.0 & 83.1 & - & 75.0 & 61.8 & 55.8 & 95.7\% \\
PruneSID & \textsc{Img.} & 58.8 & 68.5 & 54.7 & 86.5 & 1749 & 75.3 & 62.1 & - & 96.3\% \\
OccamToken & \textsc{Img.+Qry.} & \textbf{60.9} & \underline{69.1} & \textbf{58.0} & 86.3 & \textbf{1825} & \textbf{77.9} & \textbf{63.9} & \textbf{57.5} & \textbf{99.1\%} \\
\midrule

\rowcolor{gray!20} \multicolumn{11}{c}{\textit{Retain Averaged 64 Tokens ($\downarrow$ 88.9\%)}} \\
\midrule
VisionZip $\ddagger$ & \textsc{Learned} & 57.0 & 68.8 & 56.0 & 80.9 & 1756 & 74.2 & 61.5 & - & 95.1\% \\
TwigVLM & \textsc{Learned} & 58.8 & \textbf{70.0} & 55.8 & 82.7 & \underline{1760} & 75.6 & 60.4 & - & 96.0\% \\
LearnPruner & \textsc{Learned} & \underline{58.9} & 68.3 & \underline{56.6} & \textbf{86.8} & 1750 & \underline{76.0} & \underline{62.6} & \underline{55.7} & \underline{96.9\%} \\
FastV & \textsc{Fixed} & 46.1 & 51.1 & 47.8 & 48.0 & 1256 & 55.0 & 48.0 & 42.7 & 71.4\% \\
SparseVLM & \textsc{Fixed} & 52.7 & 62.2 & 51.8 & 75.1 & 1505 & 68.2 & 56.2 & 46.1 & 85.6\% \\
DivPrune & \textsc{Fixed} & 57.8 & 68.2 & 54.7 & 85.6 & 1674 & 74.1 & 59.3 & 52.3 & 93.9\% \\
DART & \textsc{Fixed} & 54.7 & 69.3 & 54.7 & 73.9 & 1705 & 71.3 & 59.5 & 54.0 & 91.9\% \\
VisPruner & \textsc{Fixed} & 55.4 & 69.1 & 55.8 & 80.4 & 1689 & 72.7 & 61.3 & 55.1 & 93.9\% \\
VisionZip & \textsc{Fixed} & 55.1 & 69.0 & 55.5 & 77.0 & 1690 & 72.4 & 60.1 & 55.4 & 93.0\% \\
PruMerge+ & \textsc{Img.} & 55.4 & \underline{69.5} & 52.0 & 75.7 & - & 71.3 & 59.6 & 52.1 & 91.3\% \\
PruneSID & \textsc{Img.} & 57.1 & 67.8 & 54.2 & 83.8 & 1733 & 73.7 & 58.8 & - & 94.0\% \\
OccamToken & \textsc{Img.+Qry.} & \textbf{59.3} & 69.0 & \textbf{57.4} & \underline{86.2} & \textbf{1801} & \textbf{77.2} & \textbf{63.2} & \textbf{56.9} & \textbf{98.1\%} \\
\midrule

\rowcolor{gray!20} \multicolumn{11}{c}{\textit{Retain Averaged 32 Tokens ($\downarrow$ 94.4\%)}} \\
\midrule
LearnPruner & \textsc{Learned} & \underline{57.2} & 68.2 & \underline{56.1} & \underline{84.5} & \underline{1672} & \underline{74.0} & \underline{60.8} & \underline{55.5} & \underline{94.8\%} \\
FastV & \textsc{Fixed} & 41.5 & 42.6 & 42.5 & 32.5 & 1090 & 43.4 & 37.8 & 33.2 & 58.5\% \\
SparseVLM & \textsc{Fixed} & 48.3 & 57.3 & 46.1 & 67.9 & 1290 & 58.6 & 51.4 & 40.6 & 76.5\% \\
DivPrune & \textsc{Fixed} & 54.9 & 68.6 & 52.9 & 81.5 & 1611 & 71.2 & 57.6 & 49.1 & 90.5\% \\
DART & \textsc{Fixed} & 52.9 & \textbf{69.3} & 52.2 & 69.1 & 1615 & 67.1 & 58.5 & 50.0 & 88.0\% \\
VisPruner & \textsc{Fixed} & 52.2 & \underline{69.2} & 53.9 & 72.7 & 1567 & 67.7 & 58.4 & 52.7 & 89.0\% \\
VisionZip & \textsc{Fixed} & 51.8 & 68.8 & 53.1 & 68.7 & 1536 & 67.1 & 57.7 & 50.3 & 87.2\% \\
PruMerge+ & \textsc{Img.} & 52.9 & 67.9 & 49.2 & 66.7 & - & 65.6 & 55.1 & 45.9 & 84.7\% \\
OccamToken & \textsc{Img.+Qry.} & \textbf{59.1} & 69.1 & \textbf{56.3} & \textbf{86.2} & \textbf{1780} & \textbf{75.5} & \textbf{62.9} & \textbf{56.2} & \textbf{97.2\%} \\
\bottomrule
\end{tabular}
}
\caption{Performance comparison on LLaVA-v1.5 7B.
\textsc{Learned}: methods with trainable pruning modules; 
\textsc{Fixed}: methods using a fixed retained-token count or ratio; 
\textsc{Img.}: image-adaptive budget allocation; 
\textsc{Img.+Qry.}: joint image- and query-adaptive budget allocation.}
\label{tab:llavav15}

\end{table}

\section{Experiments}
\label{sec:experiments}

We evaluate OccamToken on image-understanding tasks using LLaVA-v1.5 7B~\citep{liu2024improved}, LLaVA-NeXT 7B~\citep{liu2024llavanext}, and Qwen3-VL 8B~\citep{bai2025qwen3}, covering the standard 576-token setting, the high-resolution 2,880-token setting, and a non-LLaVA-family architecture. 
Following prior work~\citep{takezoelearnpruner,alvar2025divprune}, we report results on GQA~\citep{hudson2019gqa}, ScienceQA~\citep{lu2022learn}, POPE~\citep{li2023evaluating}, MME~\citep{fumme}, MMBench~\citep{liu2024mmbench}, VizWiz~\citep{gurari2018vizwiz} and RealworldQA(RQA.)~\citep{xai2024realworldqa}. 
The full-token model serves as the upper bound for each backbone. 
We report benchmark-specific scores and relative accuracy (RelAcc.), computed by normalizing performance to the corresponding full-token model. 
For adaptive methods, we report the average retained visual-token count and compare against fixed-budget baselines under matched average budgets. 
Additional results and analyses are provided in the appendix.

\begin{table}[!t]
\centering
\captionsetup{font=footnotesize}
\footnotesize
\setlength{\tabcolsep}{4pt}
\resizebox{\textwidth}{!}{%
\begin{tabular}{llc|ccccccc|c}
\toprule
Method & Venue & Params. & GQA & SQA$^{I}$ & VQA$^{T}$ & MME & VQA$^{v2}$ & MMB & POPE & RelAcc. \\
\midrule
\multicolumn{11}{c}{\cellcolor{gray!20}\textit{Upper Bound, 2,880 Tokens (100\%)}} \\
\midrule
Vanilla & - & - & 64.2 & 70.2 & 61.3 & 1842 & 81.2 & 67.9 & 86.8 & 100\% \\
\midrule

\multicolumn{11}{c}{\cellcolor{gray!20}\textit{Retain Averaged 320 Tokens ($\downarrow$ 88.9\%)}} \\
\midrule
TwigVLM     & ICCV-25  & 610M   & \underline{62.2} & \textbf{68.7} & 57.4 & \underline{1758} & \underline{79.7} & 65.0 & -- & \underline{96.3\%}\\
DivPrune    & CVPR-25  & -      & 61.1 & \underline{67.7} & 56.2 & 1721 & 77.2 & \underline{65.1} & \underline{84.7} & 95.0\% \\
VisionZip   & CVPR-25  & -      & 59.3 & 67.3 & \textbf{58.9} & 1689 & 76.2 & 63.4 & 82.1 & 94.0\% \\
DART        & EMNLP-25 & -      & 59.5 & 67.5 & 57.6 & 1710 & 75.7 & 64.2 & 81.0 & 93.8\% \\
OccamToken  & -        & -      & \textbf{63.0} & \textbf{68.7} & \underline{58.5} & \textbf{1771} & \textbf{80.0} & \textbf{65.5} & \textbf{85.7} & \textbf{97.3\%} \\
\midrule

\multicolumn{11}{c}{\cellcolor{gray!20}\textit{Retain Averaged 160 Tokens ($\downarrow$ 94.4\%)}} \\
\midrule
DivPrune    & CVPR-25  & -      & \underline{59.3} & 67.2 & 54.1 & 1614 & \underline{75.0} & \underline{63.2} & \underline{80.0} & \underline{91.7\%} \\
VisionZip   & CVPR-25  & -      & 55.5 & \underline{68.3} & \underline{56.2} & 1607 & 71.4 & 60.4 & 74.8 & 89.4\% \\
DART        & EMNLP-25 & -      & 56.8 & 67.8 & 54.9 & \underline{1700} & 72.5 & 62.0 & 75.3 & 90.6\% \\
OccamToken  & -        & -      & \textbf{61.3} & \textbf{68.5} & \textbf{57.4} & \textbf{1740} & \textbf{77.2} & \textbf{65.2} & \textbf{85.9} & \textbf{95.9\%} \\
\midrule

\multicolumn{11}{c}{\cellcolor{gray!20}\textit{Retain Averaged 40 Tokens ($\downarrow$ 98.6\%)}} \\
\midrule
VisionZip   & CVPR-25  & -      & 42.2 & 65.2 & 43.0 & 991  & 48.9 & 51.2 & 24.6 & 63.8\% \\
DivPrune    & CVPR-25  & -      & \underline{54.0} & \underline{66.8} & \underline{49.1} & \underline{1257} & \underline{67.8} & \underline{57.7} & \underline{76.4} & \underline{83.4\%} \\
OccamToken  & -        & -      & \textbf{59.5} & \textbf{68.5} & \textbf{51.8} & \textbf{1744} & \textbf{75.6} & \textbf{64.9} & \textbf{85.7} & \textbf{93.8\%} \\
\bottomrule
\end{tabular}%
}
\vspace{3pt}
\caption{Performance comparison on LLaVA-NeXT 7B under different token retention budgets.}
\label{tab:llavanext}

\end{table}

\begin{table}[!t]
\centering
\scriptsize

\makebox[\textwidth][c]{%
\begin{minipage}[t]{0.65\textwidth}
\vspace{0pt}
\centering
\setlength{\tabcolsep}{3.0pt}
\renewcommand{\arraystretch}{1.22}
\resizebox{\linewidth}{!}{%
\begin{tabular}{@{}lcccccccc|c@{}}
\toprule
Method & MME & MMB & VizWiz & POPE & GQA & RQA & VQA$^{T}$ & SQA & RelAcc. \\
\midrule
\rowcolor{gray!20}
\multicolumn{10}{c}{\textit{Upper Bound (100\%)}} \\[-1pt]
Baseline & 2375 & 84.5 & 44.8 & 88.5 & 61.9 & 71.5 & 80.1 & 94.2 & 100\% \\
\midrule
\rowcolor{gray!20}
\multicolumn{10}{c}{\textit{Retain 22.2\% ($\downarrow$77.8\%)}} \\[-1pt]
DivPrune
& 2073 & 82.7 & 42.1 & 87.4 & 57.8 & 63.4 & 72.1 & 84.8 & 92.5\% \\
\textbf{Ours}
& 2314 & 81.5 & 41.9 & 89.2 & 60.0 & 68.1 & 74.1 & 86.8 & \textbf{95.6\%} \\
\midrule
\rowcolor{gray!20}
\multicolumn{10}{c}{\textit{Retain 11.1\% ($\downarrow$88.9\%)}} \\[-1pt]
DivPrune
& 1880 & 78.7 & 38.8 & 83.7 & 54.4 & 59.2 & 65.9 & 80.3 & 86.5\% \\
\textbf{Ours}
& 2243 & 79.2 & 39.0 & 89.0 & 58.4 & 65.6 & 70.2 & 81.4 & \textbf{92.0\%} \\
\bottomrule
\end{tabular}%
}

\captionsetup{font=footnotesize,width=0.95\linewidth}
\caption{Performance comparison on Qwen3-VL 8B.}
\label{tab:qwen3}
\end{minipage}%
\hspace{0.015\textwidth}%
\begin{minipage}[t]{0.315\textwidth}
\vspace{0pt}
\centering
\scriptsize
\setlength{\tabcolsep}{1.2pt}
\renewcommand{\arraystretch}{0.96}
\resizebox{\linewidth}{!}{%
\begin{tabular}{@{}lccccc@{}}
\toprule
\textbf{Method} 
& \textbf{Tok.} 
& \textbf{Pre.} 
& \textbf{Dec.} 
& \textbf{KV} 
& \textbf{Acc.} \\
&  & \textbf{ms} & \textbf{ms} & \textbf{MB} &  \\
\midrule
v1.5 & 576 & 59 & 13.0 & 313 & 85.9 \\
VisionZip & 128 & 28 & 12.3 & 89 & 83.2 \\
\textbf{Ours} & 128 & 30 & 12.4 & 92 & \textbf{86.3} \\
VisionZip & 32 & 23 & 12.1 & 41 & 68.7 \\
\textbf{Ours} & 32 & 26 & 12.3 & 47 & \textbf{86.2} \\
\midrule
NeXT & 2880 & 164 & 16.2 & 1093 & 86.8 \\
VisionZip & 160 & 45 & 11.9 & 105 & 74.8 \\
\textbf{Ours} & 160 & 53 & 12.1 & 125 & \textbf{85.9} \\
\bottomrule
\end{tabular}%
}

\captionsetup{font=footnotesize,width=\linewidth}
\caption{Efficiency on LLaVA models.}
\label{tab:speed}
\end{minipage}%
}

\end{table}

\subsection{Main Results on Image Understanding}
\label{sec:image_main_results}

Tables~\ref{tab:llavav15} and~\ref{tab:llavanext} report the main image-understanding results on LLaVA-v1.5 7B and LLaVA-NeXT 7B. 
These two backbones represent different visual-sequence regimes: LLaVA-v1.5 uses the standard 576-token image setting, while LLaVA-NeXT processes high-resolution inputs with up to 2,880 visual tokens. 
For fair comparison with fixed-budget baselines, we match the average number of retained visual tokens for OccamToken and report both benchmark-specific scores and relative accuracy.

\textbf{Results on LLaVA-v1.5.}
Table~\ref{tab:llavav15} compares OccamToken with learned, fixed-budget, and image-adaptive pruning methods on LLaVA-v1.5 7B~\citep{liu2024improved}. 
At average budgets of 128, 64, and 32 visual tokens, OccamToken preserves 99.1\%, 98.1\%, and 97.2\% relative accuracy, respectively. 
This indicates that the method remains close to the full-token model even when the visual sequence is substantially reduced. 
Compared with training-free fixed-budget baselines~\citep{yang2025visionzip,wen2025stop}, OccamToken shows a more stable performance under aggressive compression. 
For example, at the 32-token budget, several fixed-budget methods exhibit larger drops in relative accuracy, while OccamToken maintains stronger aggregate performance. 
OccamToken is also competitive with learned pruning methods~\citep{shao2025growing,takezoelearnpruner}, despite requiring no trainable pruning module or additional training data.

\textbf{Results on LLaVA-NeXT.}
Table~\ref{tab:llavanext} evaluates OccamToken on LLaVA-NeXT 7B~\citep{liu2024llavanext}, where the original visual sequence is much longer. 
This setting is more demanding because both the prefill cost and the amount of image-dependent redundancy increase with the number of visual tokens. 
At average budgets of 320 and 160 tokens, OccamToken preserves 97.3\% and 95.9\% relative accuracy, respectively. 
Under the more aggressive 40-token setting, which retains only about 1.4\% of the original visual tokens, OccamToken still maintains 93.8\% relative accuracy. 
The compared fixed-budget methods~\citep{alvar2025divprune} degrade more noticeably in this low-budget regime, suggesting that adaptive token allocation is particularly useful when the retained visual evidence is highly constrained.

Overall, these results show that OccamToken provides a consistent accuracy--compression trade-off across both standard and long visual-sequence settings. 
Rather than assigning the same retained-token count to every sample, the register-anchored threshold allows the retained budget to vary with the current attention distribution. 
This behavior is especially beneficial under stronger compression, where preserving a small amount of task-relevant visual evidence becomes more important.

\textbf{Transfer to Another Backbone.} To examine whether OccamToken is tied to a specific LLaVA-style architecture, we further evaluate it on Qwen3-VL 8B~\citep{bai2025qwen3}. 
As shown in Table~\ref{tab:qwen3}, OccamToken improves over the compared training-free baseline under both retention ratios. 
At 22.2\% token retention, it achieves 95.6\% relative accuracy; at the stronger 11.1\% retention setting, it preserves 92.0\% relative accuracy. 
These results suggest that the register-anchored pruning rule is not tied to LLaVA-specific token interfaces and can be applied to VLMs beyond the LLaVA family.

\subsection{Ablation Study.}

\begin{figure}[t]
  \centering
  \includegraphics[width=1\textwidth]{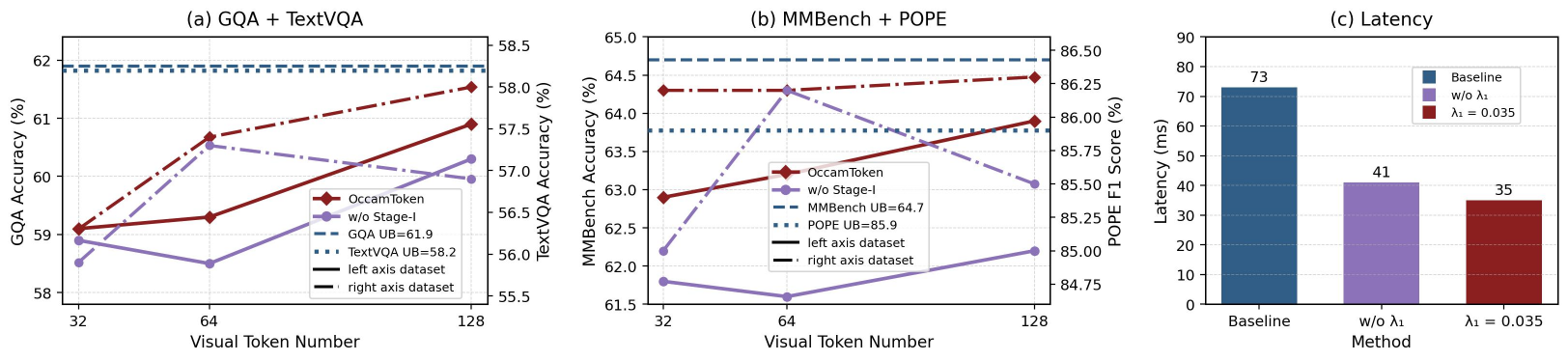}
\caption{
\textbf{Ablation and efficiency analysis.}
Full OccamToken consistently improves over the variant without Stage~I under matched token budgets, confirming the benefit of two-stage adaptive pruning. 
Stage-I also reduces latency, yielding a better accuracy--efficiency trade-off.
}
\label{fig:ablation_efficiency}

\end{figure}

We ablate the key components of OccamToken to examine whether the two-stage design is necessary. 
The comparison between OccamToken and the variant without Stage~I is particularly informative. 
Although both variants perform query-adaptive pruning in the language model, removing Stage~I consistently reduces relative accuracy under matched average token budgets. 
This shows that Stage~I is not merely an efficiency shortcut; it removes image-level redundancy before the visual tokens enter LLM computation and provides a cleaner candidate set for subsequent query-adaptive pruning.

\subsection{Efficiency Analysis}

We further evaluate whether OccamToken improves inference efficiency while preserving visual understanding. 
Since visual tokens dominate the multimodal prefix, reducing the retained visual sequence directly lowers the prefill cost and KV-cache memory. 
On LLaVA-v1.5 7B, OccamToken reduces the visual sequence from 576 tokens to 32 tokens, decreasing the KV cache from 313 MB to 47 MB while maintaining 86.2 POPE accuracy. 
On LLaVA-NeXT 7B, it reduces 2,880 visual tokens to 160 tokens, lowering the KV cache from 1093 MB to 125 MB while preserving 85.9 POPE accuracy.
Compared with simpler fixed-budget methods such as VisionZip~\citep{yang2025visionzip}, OccamToken is not always the fastest method at the same final token count because it introduces lightweight two-stage adaptive scoring. 
However, this small overhead leads to substantially better accuracy preservation, particularly under aggressive compression. 
Therefore, OccamToken should be understood not as a latency-only method, but as a practical accuracy--efficiency trade-off: it significantly reduces prefill and memory costs while retaining reliable visual evidence.

\FloatBarrier

\section{Conclusion}

This paper presents OccamToken, a training-free framework for efficient VLM inference through register-anchored visual token pruning. Instead of relying on fixed top-$K$ selection, OccamToken uses a test-time register token as a dynamic reference to mitigate attention-sink distortion and adapt the pruning threshold to each input. Its two-stage design further combines image-level redundancy removal with query-conditioned relevance selection, enabling adaptive token allocation without additional training or trainable pruning modules. Experiments across LLaVA-v1.5, LLaVA-NeXT and Qwen3-VL show that OccamToken consistently improves the accuracy--efficiency trade-off. In particular, it pushes training-free visual token compression into the 1.4\% retention regime on LLaVA-NeXT while preserving over 93\% of the full-token accuracy. These results suggest that reference-based adaptive pruning is a simple and effective direction for scalable multimodal inference.

\textbf{Limitations.}
OccamToken is training-free at the pruning stage, but it relies on constructing a reliable test-time register token for the target VLM.
While this is effective for the Transformer-based backbones evaluated in this work, architectures with different feed-forward blocks, attention interfaces, or vision--language projection designs may require additional adaptation.
Extending register construction and reference-based pruning to broader multimodal architectures is a natural direction for future work.

\FloatBarrier

\bibliographystyle{assets/plainnat}
\bibliography{paper}

\newpage
\beginappendix

The appendix is organized as follows. Appendix~\ref{sec:setup} provides implementation details, including parameter settings, register construction, pruning-layer selection, architecture-specific instantiations, efficiency measurement, and component ablations. Appendix~\ref{sec:benchmarks} describes the image and video benchmarks used in our evaluation, together with their evaluation protocols and metrics. Appendix~\ref{sec:more_results} reports additional experimental results, including video understanding, Mini-Gemini, larger LLaVA backbones, and transfer experiments on non-LLaVA architectures. Appendix~\ref{sec:bimpact} discusses the broader impact of efficient VLM inference.

\section{Implementation Details}
\label{sec:setup}

\begin{figure}[t]
\centering
\includegraphics[width=1\textwidth]{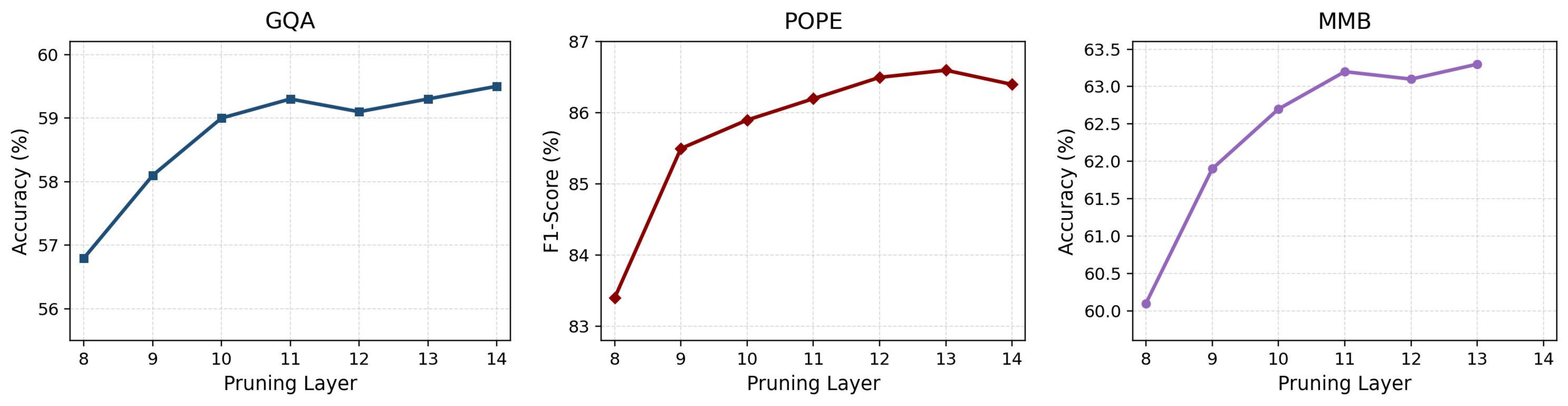}
\caption{\textbf{Effect of the Stage~II pruning layer on LLaVA-v1.5.}
We evaluate OccamToken by applying query-adaptive pruning at different LLM decoder layers under the same token budget.
Across GQA, POPE, and MMB, performance improves rapidly from shallow layers and becomes largely stable around layer 11.
Considering both accuracy and efficiency, we select layer 11 as the default Stage-II pruning layer for LLaVA-v1.5.}
\label{fig:layerprune}
\end{figure}

\begin{table}[t]
\centering
\small
\setlength{\tabcolsep}{6pt}
\resizebox{\columnwidth}{!}{%
\begin{tabular}{lcccccc}
\toprule
Setting 
& $\lambda_1$ 
& $\lambda_2$
& $N_1$ 
& $N_2$ 
& Speedup 
& POPE \\
\midrule
w/o Stage I 
& 0
& 1 
& 2880 
& 160.8 
& 1.00$\times$ 
& 85.7 \\
Mild 
& 0.01 
& 1.08
& 1274 
& 163.9 
& 1.21$\times$ 
& 86.3 \\
Moderate 
& 0.025
& 0.62 
& 833 
& 157.4 &
1.37$\times$ &
86.0 \\
Aggressive 
& 0.05 
& 1.3
& 470 
& 161.3 
& 1.54$\times$ 
& 85.6 \\
Aggressive 
& \textbf{0.045 }
& 1.3
& 523 
& 159.3 
& \textbf{1.51$\times$ }
& \textbf{85.9} \\
\bottomrule
\end{tabular}%
}
\vspace{10pt}
\caption{
Stage~I ablation and $\lambda_1$ sensitivity on POPE with LLaVA-NeXT.
We adjust $\lambda_2$ to keep the final retained token budget comparable across settings.
$N_1$ and $N_2$ denote the average number of visual tokens retained after Stage~I and Stage~II.
Speedup is computed against the setting without Stage~I pruning.
}
\label{tab:lamda1cc}
\end{table}

\paragraph{Parameter settings.}
All experiments are conducted on a single NVIDIA RTX PRO 6000 GPU.
We select the pruning layer and Stage-I coefficient according to the empirical accuracy--efficiency trade-off shown in Figure~\ref{fig:layerprune} and Table~\ref{tab:lamda1cc}.

For Stage-II pruning, Figure~\ref{fig:layerprune} shows that the pruning layer has a clear effect on downstream performance.
When query-adaptive pruning is applied at shallow decoder layers, the performance is consistently lower on GQA, POPE, and MMB.
This suggests that early LLM layers have not yet formed sufficiently reliable text-to-vision alignment, making their attention signals less suitable for query-conditioned token selection.
As the pruning layer moves deeper, performance improves rapidly and becomes largely stable around layer 11.
After this point, applying pruning at later layers brings only marginal accuracy gains, but reduces the efficiency benefit because more decoder layers have already processed the full visual sequence.
Therefore, we choose layer 11 as the default Stage~II pruning layer for LLaVA-series models, which provides a practical balance between reliable query-aware attention and early enough token removal.
For Qwen-series models, we use layer 10 following the same principle: pruning is applied at an intermediate decoder layer where cross-modal attention is sufficiently formed while substantial later computation can still benefit from the reduced visual prefix.

For Stage~I pruning, we set $\lambda_1$ according to the visual-token scale of each backbone.
For LLaVA-v1.5, whose visual prefix contains 576 tokens, we use $\lambda_1$ in the range of 0.01--0.02, which keeps about 200--300 candidate tokens before Stage~II pruning.
For LLaVA-NeXT, whose visual prefix contains 2,880 tokens, stronger Stage~I pruning is necessary to remove the much larger amount of image-level redundancy, and we use $\lambda_1$ in the range of 0.04--0.05.

Table~\ref{tab:lamda1cc} provides a detailed sensitivity analysis on LLaVA-NeXT.
Without Stage~I, Stage~II must operate over all 2,880 visual tokens and obtains 85.7 POPE with baseline speed.
A mild Stage~I setting already reduces the candidate set to 1,274 tokens and improves the speed to 1.21$\times$, while slightly increasing POPE to 86.3.
A moderate setting with $\lambda_1=0.025$ further reduces the candidate set to 833 tokens and reaches 1.37$\times$ speedup with 86.0 POPE.
Among the stronger settings, we choose $\lambda_1=0.045$ as the default for LLaVA-NeXT.
This setting keeps 523 candidate tokens after Stage~I, achieves a 1.51$\times$ speedup, and maintains 85.9 POPE.
Compared with $\lambda_1=0.025$, it preserves the same POPE score while improving speed from 1.37$\times$ to 1.51$\times$.
Compared with the more aggressive $\lambda_1=0.05$, it retains nearly the same speedup, 1.51$\times$ versus 1.54$\times$, but improves POPE from 85.6 to 85.9.
Thus, $\lambda_1=0.045$ lies near the practical Pareto frontier: it captures almost the efficiency gain of aggressive pruning while avoiding the accuracy degradation caused by removing too many candidates too early.

This parameter choice also matches the intended role of Stage~I.
Stage~I is query-agnostic, so it should remove broad image-level redundancy rather than make the final evidence-selection decision.
If Stage-I is too weak, many redundant tokens remain and the speedup is limited.
If Stage~I is too aggressive, potentially query-relevant evidence may be removed before Stage~II can evaluate it.
The selected $\lambda_1=0.045$ removes over 80\% of the original visual tokens before Stage~II, keeps the final budget comparable, and still maintains accuracy above the no-Stage~I baseline.
This makes it a robust default for high-resolution LLaVA-NeXT experiments.

We construct one register token using the top-10 register-related sparse neurons.
Since these neurons are stable across samples, we identify them from a small calibration subset and cache the neuron indices for reuse.
OccamToken does not enforce a hard token budget.
Instead, we use 50--100 calibration samples to estimate the relationship between pruning coefficients and the average retained-token count, and then report the average budget over the full evaluation set.
In practice, the results are stable under small coefficient variations, making budget matching simple and reliable.

\paragraph{Register construction.}
For Stage~I, we construct the test-time register token following Jiang et al.~\cite{jiangvision}.
Let $\mathcal{V}=\{v_i\}_{i=1}^{N}$ denote the visual patch tokens produced by the vision encoder, and let $h_{i,k}$ denote the activation of token $v_i$ on channel $k$ in the corresponding MLP layer.
Given the register-related neuron set $\mathcal{K}_{\mathrm{reg}}$, we redirect sink-like channel activations from ordinary visual tokens to the appended register token $r$:
\begin{equation}
h_{r,k} \leftarrow \max_{v_i\in\mathcal{V}} h_{i,k}, 
\qquad
h_{i,k} \leftarrow 0,\ \forall v_i\in\mathcal{V},\quad
k\in\mathcal{K}_{\mathrm{reg}}.
\label{eq:register_redirect_appendix}
\end{equation}
The register token is then inserted into the visual-token sequence, forming $\widetilde{\mathcal{V}}=\mathcal{V}\cup\{r\}$, and participates in the same attention computation as ordinary patch tokens.
Unless otherwise specified, the register token is always preserved after pruning, since it serves as the reference anchor for both image-adaptive and query-adaptive pruning.

\paragraph{Stage-I pruning.}
Stage~I is applied at the output of the vision encoder before visual tokens are projected into the language-model embedding space.
For VLMs whose vision encoder provides a \texttt{[CLS]} token, such as LLaVA-v1.5 and LLaVA-NeXT, we use the \texttt{[CLS]} token as the image-level evaluator.
Let $A^{\mathrm{ve}}$ denote the attention map in the vision encoder after register construction, averaged over attention heads when multi-head attention is used.
We compute the Stage-I visual-token score and register reference score as
\begin{equation}
s_1(v_i)=A^{\mathrm{ve}}_{\mathrm{cls}\rightarrow v_i},
\qquad
s_1(r)=A^{\mathrm{ve}}_{\mathrm{cls}\rightarrow r}.
\label{eq:stage1_cls_appendix}
\end{equation}
The Stage-I pruning threshold is defined by the scaled register score,
\begin{equation}
\tau_1 = \lambda_1 s_1(r),
\label{eq:stage1_threshold_appendix}
\end{equation}
and the retained token set is
\begin{equation}
\mathcal{S}_1
=
\{v_i\in\mathcal{V}\mid s_1(v_i)\geq \tau_1\}\cup\{r\}.
\label{eq:stage1_set_appendix}
\end{equation}

For architectures without an explicit \texttt{[CLS]} token, we use the mutual-scoring variant described in Section~\ref{sec:stage1}.
In this case, token importance is estimated by the average attention received from visual and register tokens:
\begin{equation}
\begin{aligned}
s^{\mathrm{mut}}_1(v_i)
&=
\frac{1}{|\mathcal{V}|+1}
\sum_{u\in\mathcal{V}\cup\{r\}}
A^{\mathrm{ve}}_{u\rightarrow v_i},
\\
s^{\mathrm{mut}}_1(r)
&=
\frac{1}{|\mathcal{V}|+1}
\sum_{u\in\mathcal{V}\cup\{r\}}
A^{\mathrm{ve}}_{u\rightarrow r}.
\end{aligned}
\label{eq:stage1_mutual_appendix}
\end{equation}
The same pruning rule in Eq.~\eqref{eq:stage1_set_appendix} is then applied by replacing $s_1(\cdot)$ with $s^{\mathrm{mut}}_1(\cdot)$.

\paragraph{Stage-II pruning.}
Stage~II is applied inside the language model to further remove query-irrelevant visual tokens.
After Stage~I, the retained visual tokens and the register token are projected into the language embedding space:
\begin{equation}
\mathbf{Z}_1 = p(\mathcal{S}_1),
\qquad
\mathbf{X}^{(0)}=[\mathbf{Z}_1;\mathbf{T}],
\label{eq:project_appendix}
\end{equation}
where $p(\cdot)$ denotes the visual projector and $\mathbf{T}=\{t_j\}_{j=1}^{M}$ denotes the embedded text query.
We run the language model until the pruning layer $l$ and extract the text-to-visual/register attention scores.
Following our main setting, Stage~II is applied at the 11th decoder layer.

For each surviving visual token, we compute its query-conditioned score by the maximum attention received from all text tokens:
\begin{equation}
s_2(v_i)
=
\max_{j=1,\ldots,M}
A^{(l)}_{t_j\rightarrow v_i},
\qquad
v_i\in\mathcal{S}_1\setminus\{r\}.
\label{eq:stage2_visual_score_appendix}
\end{equation}
For the register token, we use the mean text-to-register attention as a stable query-conditioned reference:
\begin{equation}
s_2(r)
=
\frac{1}{M}
\sum_{j=1}^{M}
A^{(l)}_{t_j\rightarrow r}.
\label{eq:stage2_register_score_appendix}
\end{equation}
The Stage~II threshold and final retained set are therefore
\begin{equation}
\begin{aligned}
\tau_2 = \lambda_2 s_2(r),
\\
\mathcal{S}_2
&=
\{v_i\in\mathcal{S}_1\setminus\{r\}\mid s_2(v_i)\geq \tau_2\}
\\
&\quad \cup \{r\}.
\end{aligned}
\label{eq:stage2_set_appendix}
\end{equation}
The coefficient $\lambda_2$ controls the overall compression strength and is adjusted only to match the target average token budget used for comparison with fixed-budget baselines.

\paragraph{Dynamic budgets and fair comparison.}
Unlike fixed top-$K$ pruning, OccamToken does not enforce the same number of retained tokens for every input.
For an evaluation set $\mathcal{D}$, the reported retained token count is the dataset-level average
\begin{equation}
\overline{K}
=
\frac{1}{|\mathcal{D}|}
\sum_{(I,T)\in\mathcal{D}}
|\mathcal{S}_2(I,T)|.
\label{eq:average_budget_appendix}
\end{equation}
When comparing against a fixed-budget baseline with target budget $K_{\mathrm{tar}}$, we select $\lambda_2$ such that the average retained token count matches the target budget:
\begin{equation}
\lambda_2^{\star}
=
\arg\min_{\lambda_2}
\left|
\frac{1}{|\mathcal{D}|}
\sum_{(I,T)\in\mathcal{D}}
|\mathcal{S}_2(I,T;\lambda_2)|
-
K_{\mathrm{tar}}
\right|.
\label{eq:lambda2_budget_matching_appendix}
\end{equation}
This protocol ensures that performance differences are attributed to the pruning criterion rather than to retaining more visual tokens on average.
For LLaVA-v1.5, we match the average budgets of 128/64/32 tokens.
For LLaVA-NeXT, we match the average budgets of 640/320/160/40 tokens.
Because OccamToken uses a dynamic pruning boundary, individual samples may retain more or fewer tokens than the target budget depending on their image-level redundancy and query-conditioned evidence demand.

\paragraph{Architecture-specific instantiations.}
For LLaVA-v1.5 and LLaVA-NeXT, Stage~I uses \texttt{[CLS]}-to-register and \texttt{[CLS]}-to-visual attention in the vision encoder.
For Qwen2-VL, whose visual encoder does not expose an explicit \texttt{[CLS]} token, we instantiate Stage~I with the mutual-scoring scheme in Eq.~\eqref{eq:stage1_mutual_appendix}.
For Video-LLaVA, we treat visual tokens from all frames as a unified visual-token set:
\begin{equation}
\mathcal{V}_{\mathrm{video}}
=
\bigcup_{f=1}^{F}
\mathcal{V}^{(f)},
\label{eq:video_tokens_appendix}
\end{equation}
where $\mathcal{V}^{(f)}$ denotes the visual tokens extracted from the $f$-th frame.
We then apply the same two-stage pruning procedure without additional temporal modules.
This keeps OccamToken training-free and plug-and-play across image and video VLMs.

\paragraph{Efficiency measurement.}
All experiments are conducted on a single NVIDIA Pro 6000 GPU.
For efficiency evaluation, we reportprefill latency, decode time, and KV-cache memory.
Let $N$ denote the original number of visual tokens and $\overline{K}$ denote the average number of retained visual tokens after pruning.
Since the language-model prefill cost grows with the multimodal sequence length, reducing the visual prefix changes the effective prefill length from $N+M$ to approximately $\overline{K}+M$, where $M$ is the text length.
The KV-cache memory also scales linearly with the retained sequence length:
\begin{equation}
\mathrm{Mem}_{\mathrm{KV}}
\propto
2 L H d_h \cdot (\overline{K}+M),
\label{eq:kv_memory_appendix}
\end{equation}
where $L$ is the number of decoder layers, $H$ is the number of attention heads, and $d_h$ is the head dimension.
Prefill latency is averaged over the evaluation samples under the same decoding configuration.
Since OccamToken only introduces register construction and lightweight threshold comparisons, the measured speedup mainly comes from reducing the number of visual tokens processed by the language model.

\paragraph{Compact component ablation.}
Table~\ref{tab:qwen_compact_component_ablation} summarizes a compact component ablation on Qwen3-VL 8B under matched average token retention ratios. 
OccamToken achieves the highest RelAcc. at both 22.2\% and 11.1\% retention, obtaining 95.6\% and 92.0\%, respectively. 
Compared with the fixed-budget variant, OccamToken improves RelAcc. by 2.5 and 3.3 points, indicating that adaptive token allocation is beneficial when the retained visual budget is constrained. 
Removing Stage-I leads to smaller drops of 0.4 and 0.2 RelAcc., suggesting that image-level redundancy pruning provides a modest but consistent complementary benefit before query-adaptive pruning. 
Replacing the register-anchored threshold with a mean-based threshold reduces RelAcc. by 0.7 and 1.3 points, showing that the register score is a more effective pruning reference than a generic mean statistic. 
Furthermore, removing register construction from the mean-threshold variant further decreases RelAcc. from 94.9 to 94.3 at 22.2\% retention and from 90.7 to 90.2 at 11.1\% retention, suggesting that test-time register construction also contributes to more stable token selection. 
Overall, these results support the combined roles of adaptive allocation, register construction, register-anchored thresholding, and Stage-I pruning in OccamToken.

\section{Evaluation Benchmarks}
\label{sec:benchmarks}

\begin{table}[t]
\centering
\captionsetup{font=small}
\small
\setlength{\tabcolsep}{5.3pt}
\resizebox{\columnwidth}{!}{%
\begin{tabular}{lcccccc}
\toprule
\multirow{2}{*}{\textbf{Variant}}
& \multicolumn{4}{c}{\textbf{Design Choice}}
& \multicolumn{2}{c}{\textbf{RelAcc. (\%)}} \\
\cmidrule(lr){2-5} \cmidrule(lr){6-7}
& \textbf{Fixed Budget}
& \textbf{Build Reg.}
& \textbf{Reg. Threshold}
& \textbf{Stage-I}
& \textbf{22.2\%}
& \textbf{11.1\%} \\
\midrule
OccamToken
& $\times$ & \checkmark & \checkmark & \checkmark
& \textbf{95.6} & \textbf{92.0} \\

Fixed budget
& \checkmark & \checkmark & $\times$ & \checkmark
& 93.1 & 88.7 \\

w/o Stage-I
& $\times$ & \checkmark & \checkmark & $\times$
& 95.2 & 91.8 \\

Mean threshold
& $\times$ & \checkmark & $\times$ & \checkmark
& 94.9 & 90.7 \\

Mean w/o reg.
& $\times$ & $\times$ & $\times$ & \checkmark
& 94.3 & 90.2 \\

\bottomrule
\end{tabular}%
}
\vspace{3pt}
\caption{
\textbf{Compact component ablation on Qwen3-VL 8B.}
We report RelAcc. under matched average token retention ratios and ablate fixed budgeting, register construction, register-anchored thresholding, and Stage-I pruning.
}
\label{tab:qwen_compact_component_ablation}
\end{table}

We evaluate OccamToken on a diverse collection of image and video understanding benchmarks to assess its accuracy--efficiency trade-off under different visual reasoning scenarios.
The image benchmarks cover general visual question answering, compositional reasoning, OCR-oriented understanding, object hallucination, multilingual evaluation, expert-level multimodal reasoning, and robustness to real-world low-quality images.
The video benchmarks further test whether the proposed pruning strategy can handle temporally redundant visual tokens while preserving evidence for spatio-temporal reasoning.
Overall, our evaluation includes 11 image understanding benchmarks and 4 video understanding benchmarks.

Unless otherwise specified, we follow the standard evaluation protocol of each benchmark and report the commonly used metric in prior VLM evaluation.
For multiple-choice and open-ended image benchmarks, we report accuracy or the official benchmark score.
For MME, we report the total score following its original protocol.
For video question-answering benchmarks, we follow the evaluation setting used by Video-LLaVA and related video VLM baselines.
All results are computed using the same decoding configuration within each model family, so performance differences mainly reflect the effect of visual token pruning.

\subsection{Image Understanding Benchmarks}

\textbf{GQA}~\cite{hudson2019gqa} is a large-scale benchmark for real-world visual reasoning and compositional question answering.
It contains over 22 million questions generated from scene graphs, covering object attributes, spatial relationships, relational reasoning, and multi-step compositional understanding.
Compared with conventional VQA benchmarks, GQA places stronger emphasis on structured reasoning and grounding, making it useful for evaluating whether token pruning preserves fine-grained relational evidence.
We report accuracy on the test-dev balanced split.

\textbf{MMBench (MMB)}~\cite{liu2024mmbench} is an objective benchmark for evaluating general multimodal capabilities of vision-language models.
It organizes evaluation into a three-level ability hierarchy, including perception and reasoning categories, intermediate ability groups, and fine-grained ability dimensions.
MMBench contains approximately 3,000 multiple-choice questions and uses CircularEval together with ChatGPT-based answer matching.
\textbf{MMBench-CN (MMB$^{\text{CN}}$)} is the Chinese-language variant of MMBench and follows the same evaluation protocol, allowing us to examine multilingual multimodal understanding.

\textbf{MME}~\citep{fumme} evaluates both perceptual and cognitive abilities of multimodal large language models.
It covers 14 subtasks, including object existence, count, position, color, scene understanding, commonsense reasoning, numerical calculation, text translation, and code reasoning.
MME uses manually constructed instruction-answer pairs with a concise ``Yes/No'' format to reduce ambiguity in answer matching.
We report the total score, defined as the sum of accuracy and accuracy$^{+}$ across all subtasks, with a maximum score of 2800.

\textbf{POPE}~\cite{li2023evaluating} evaluates object hallucination by asking binary questions about whether a specific object exists in the image.
It constructs questions under three sampling strategies: random, popular, and adversarial.
This benchmark is particularly relevant for visual token pruning, since overly aggressive pruning may remove small or less salient objects and increase hallucination.
We report accuracy averaged across the three sampling strategies.

\textbf{ScienceQA-IMG (SQA$^{\text{I}}$)}~\cite{lu2022learn} is the image subset of the ScienceQA benchmark.
It covers diverse science topics across natural science, language science, and social science, and often requires multi-step reasoning over both visual information and textual context.
We use the image subset and report accuracy.

\textbf{TextVQA (VQA$^{\text{T}}$)}~\cite{singh2019towards} focuses on visual question answering that requires reading and reasoning about text in images.
It contains 45,336 questions on 28,408 images from the OpenImages dataset.
Since answers often depend on signs, labels, documents, and other text-rich regions, TextVQA is important for evaluating whether pruning preserves small but semantically crucial visual regions.
We report accuracy on the validation split.

\textbf{VQAv2 (VQA$^{\text{v2}}$)}~\cite{goyal2017making} is a large-scale open-ended visual question answering benchmark.
It contains 265,016 images paired with over 1.1 million questions, with 10 human-provided answers for each question.
VQAv2 reduces language priors by pairing similar questions with different images that lead to different answers.
We report accuracy on the test-dev split.

\textbf{MMMU}~\cite{yue2024mmmu} is a massive multi-discipline multimodal understanding benchmark containing 11,500 college-level questions across 30 subjects and six core disciplines.
It includes visually complex inputs such as charts, diagrams, tables, maps, chemical structures, and scientific figures.
MMMU is therefore a challenging testbed for evaluating whether token pruning preserves fine-grained, domain-specific evidence required for expert-level reasoning.
We report accuracy on the validation split.

\textbf{SEED-Bench (SEED$^{\text{I}}$)}~\cite{li2023seed} evaluates multimodal large language models across diverse visual understanding abilities.
It covers scene understanding, instance identity, attributes, location, counting, spatial relation, visual reasoning, text understanding, and action recognition.
The benchmark contains 19,000 multiple-choice questions with human annotations.
We use the image subset, denoted as SEED$^{\text{I}}$, and report accuracy.
\begin{table}[t]
\centering
\small
\setlength{\tabcolsep}{6pt}
\renewcommand{\arraystretch}{1.1}
\resizebox{0.82\textwidth}{!}{%
\begin{tabular}{l|cccccc|c}
\toprule
\textbf{Method} & \textbf{GQA} & \textbf{SQA$^{I}$} & \textbf{VQA$^{T}$} & \textbf{MME} & \textbf{POPE} & \textbf{MMB} & \textbf{Avg} \\
\midrule
\rowcolor{gray!25}
\multicolumn{8}{c}{\textit{Upper Bound, 576 Tokens (100\%)}} \\
Vanilla & 62.4 & 70.7 & 65.2 & 1841 & 85.8 & 69.3 & 100\% \\
\rowcolor{gray!25}
\multicolumn{8}{c}{\textit{Retain Averaged 192 Tokens (\textcolor{blue}{$\downarrow$ 66.7\%})}} \\
VisionZip            & 60.3          & 70.7          & 63.4          & 1846          & 82.3          & 68.9          & 98.2\% \\
VisionZip $\ddagger$ & \textbf{61.6} & 70.2          & 63.6          & 1804          & \textbf{85.5} & 67.2          & 98.4\% \\
Mutual               & 61.4          & \textbf{71.0} & \textbf{64.7} & 1837          & 84.9          & 69.1          & \textbf{99.4\%} \\
CLS                  & 61.4          & 70.8          & 64.5          & \textbf{1849} & 84.0          & \textbf{69.3} & 99.3\% \\
\rowcolor{gray!25}
\multicolumn{8}{c}{\textit{Retain Averaged 128 Tokens (\textcolor{blue}{$\downarrow$ 77.8\%})}} \\
VisionZip            & 58.7          & 70.0          & 61.3          & 1841          & 78.5          & 68.1          & 96.1\% \\
VisionZip $\ddagger$ & 60.0          & 70.1          & 61.6          & 1810          & 83.2          & 67.0          & 97.0\% \\
Mutual               & 60.7          & \textbf{70.6} & \textbf{64.3} & 1833          & 83.0          & 68.0          & 98.4\% \\
CLS                  & \textbf{60.8} & 70.3          & 64.1          & \textbf{1842} & \textbf{83.5} & \textbf{68.5} & \textbf{98.6\%} \\
\rowcolor{gray!25}
\multicolumn{8}{c}{\textit{Retain Averaged 64 Tokens (\textcolor{blue}{$\downarrow$ 88.9\%})}} \\
VisionZip            & 55.8          & 70.7          & 59.1          & 1737          & 69.6          & 65.9          & 91.8\% \\
VisionZip $\ddagger$ & 57.7          & \textbf{71.0} & 60.1          & 1779          & 80.0          & \textbf{66.3} & 95.1\% \\
Mutual               & \textbf{59.8} & 70.6          & \textbf{62.1} & \textbf{1803} & 82.8          & 66.2          & \textbf{96.8\%} \\
CLS                  & 59.6          & 70.5          & 61.5          & 1798          & \textbf{83.2} & \textbf{66.3} & 96.6\% \\
\bottomrule
\end{tabular}%
}
\vspace{10pt}
\caption{
Performance comparison on Mini-Gemini under different average token budgets.
We evaluate two Stage-I instantiations of OccamToken: \textbf{Mutual}, which uses mutual attention among visual/register tokens when no dedicated evaluator is assumed, and \textbf{CLS}, which uses \texttt{[CLS]}-based attention scoring.
Both variants consistently outperform VisionZip and VisionZip$\ddagger$ across compression levels, showing that register-anchored thresholding generalizes beyond LLaVA-style architectures and is not tied to a specific scoring mechanism.
}
\label{tab:gmm}
\end{table}

\textbf{VizWiz}~\cite{gurari2018vizwiz} consists of over 31,000 visual questions collected from visually impaired users.
The images are captured in real-world conditions and often contain challenging artifacts such as blur, poor framing, occlusion, low resolution, or irrelevant visual content.
Each question is paired with 10 crowdsourced answers, and the benchmark includes an ``unanswerable'' option when the image does not provide sufficient evidence.
VizWiz is useful for evaluating the robustness of token pruning under naturally noisy and imperfect visual inputs.
We report accuracy on the test split.

\subsection{Video Understanding Benchmarks}

Video understanding introduces additional challenges for visual token pruning.
Compared with static images, videos contain a much larger number of visual tokens due to the temporal dimension, and many adjacent frames may be highly redundant.
At the same time, important evidence may appear only in a short temporal segment or depend on motion patterns across frames.
Therefore, video benchmarks provide a natural testbed for evaluating whether OccamToken can remove temporal redundancy while preserving task-relevant spatio-temporal evidence.

\textbf{TGIF-QA}~\cite{jang2017tgif} extends visual question answering to short animated GIF videos.
It contains approximately 165,000 question-answer pairs and introduces several spatio-temporal reasoning tasks, including repetition counting, repeating action identification, and state transition detection, in addition to frame-level question answering.
These tasks require models to capture temporal dynamics rather than relying only on static appearance.
We report accuracy and GPT-evaluated scores following the Video-ChatGPT evaluation protocol.

\textbf{MSVD-QA}~\cite{xu2017video} is built on the Microsoft Research Video Description dataset.
It contains 1,970 video clips and approximately 50,500 open-ended question-answer pairs.
The questions span common types such as what, who, how, when, and where, covering a wide range of visual entities, actions, and scene-level semantics.
MSVD-QA is widely used for evaluating video question answering and provides a standard benchmark for measuring whether compressed video tokens preserve essential semantic information.
We report accuracy and GPT-evaluated scores.

\textbf{MSRVTT-QA}~\cite{xu2017video} is constructed from the MSR-VTT video dataset and contains 10,000 video clips with 243,000 question-answer pairs.
Compared with MSVD-QA, MSRVTT-QA is larger and contains more diverse real-world video content.
The questions also span what, who, how, when, and where types, requiring models to integrate visual appearance, temporal context, and language understanding.
This benchmark is useful for evaluating the scalability of token pruning under diverse video inputs.
We report accuracy and GPT-evaluated scores.

\textbf{ActivityNet-QA}~\cite{yu2019activitynet} contains 58,000 human-annotated question-answer pairs on 5,800 videos from the ActivityNet dataset.
The videos usually depict complex human activities and often require longer-range temporal reasoning.
The questions cover motion, spatial relationships, temporal relationships, and event-level understanding.
ActivityNet-QA is particularly challenging because relevant evidence may be distributed across long video segments rather than concentrated in a single frame.
We report accuracy and GPT-evaluated scores.

\section{More Experiments Results}
\label{sec:more_results}
\subsection{Additional Results on Video Understanding}

\begin{table}[t]
\centering
\captionsetup{font=footnotesize}
\small
\setlength{\tabcolsep}{8pt}
\renewcommand{\arraystretch}{1.05}
\resizebox{\columnwidth}{!}{
\begin{tabular}{lccccc}
\toprule
\textbf{Method} & \textbf{TGIF} & \textbf{MSVD} & \textbf{MSRVTT} & \textbf{ActNet} & \textbf{Rel.} \\
\midrule
Video-LLaVA & 47.1 & 69.8 & 56.7 & 43.1 & 100\% \\
\midrule
SparseVLM  & 44.7 & 68.2 & 31.0 & 42.6 & 86.5\% \\
VisionZip  & 42.4 & 63.5 & 52.1 & 43.0 & 93.2\% \\
Ours       & \textbf{44.9} & \textbf{66.7} & \textbf{56.1} & \textbf{43.5} & \textbf{97.7\%} \\
\bottomrule
\end{tabular}
}
\vspace{2pt}
\caption{
\textbf{Additional comparison on video understanding benchmarks.}
OccamToken preserves 97.7\% relative accuracy on Video-LLaVA and outperforms prior pruning methods across the averaged relative score.
}
\label{tab:videollava}

\end{table}

\paragraph{Video understanding.}
We additionally evaluate OccamToken on Video-LLaVA~\citep{lin2024video} to examine its applicability beyond image-only inputs.
As shown in Table~\ref{tab:videollava}, OccamToken preserves 97.7\% of the full-token Video-LLaVA performance, outperforming SparseVLM and VisionZip by 11.2 and 4.5 relative-accuracy points, respectively.
Notably, OccamToken remains close to the full-token baseline on MSRVTT and slightly improves the score on ActivityNet.
These results provide additional evidence that register-anchored pruning can remove redundant visual tokens while retaining the evidence needed for video understanding.

\subsection{Results on Mini-Gemini}

\textbf{Mini-Gemini.}
Table~\ref{tab:gmm} reports additional results on Mini-Gemini, a VLM equipped with a dual vision encoder architecture and an additional high-resolution encoder for fine-grained visual understanding.
This architecture differs substantially from the LLaVA family, and therefore provides a useful testbed for evaluating whether OccamToken is tied to a specific vision encoder design or a particular scoring token.
On Mini-Gemini, we evaluate two Stage-I instantiations: the \texttt{[CLS]}-based scoring scheme and the mutual-scoring scheme described in Section~\ref{sec:stage1}.

The results show that both scoring schemes effectively support register-anchored dynamic thresholding.
At the 192-token setting, the mutual-scoring variant achieves 99.4\% relative accuracy, while the \texttt{[CLS]}-based variant achieves 99.3\%, both preserving almost all of the full-token performance.
At the 128-token setting, the \texttt{[CLS]} variant obtains the best average result with 98.6\% relative accuracy, while the mutual-scoring variant remains close at 98.4\%.
Even under the more aggressive 64-token setting, both variants remain robust, achieving 96.8\% and 96.6\% relative accuracy for mutual scoring and \texttt{[CLS]} scoring, respectively.
These results indicate that the effectiveness of OccamToken does not come from a specific choice of evaluator, but from the register-anchored relative comparison principle.

Compared with VisionZip and VisionZip$\ddagger$, OccamToken variants consistently provide a stronger accuracy--efficiency trade-off, especially under tighter token budgets.
For example, when retaining only 64 tokens on average, the mutual-scoring variant improves relative accuracy from 91.8\% with VisionZip and 95.1\% with VisionZip$\ddagger$ to 96.8\%.
The gains are particularly visible on benchmarks requiring fine-grained or reliable visual evidence, such as GQA, VQA$^{T}$, MME, and POPE.
This suggests that register-anchored thresholding is better able to preserve task-relevant tokens when the compression ratio becomes high.
Overall, the Mini-Gemini results further support the architecture generality of OccamToken: the same reference-adaptive pruning principle remains effective beyond LLaVA-style VLMs and can be instantiated with either \texttt{[CLS]}-based or mutual attention scoring.

\begin{table}[t]
\centering
\small
\setlength{\tabcolsep}{5pt}
\resizebox{0.95\textwidth}{!}{%
\begin{tabular}{l|cccccccc|c}
\toprule
Method & GQA & MMB & MME & POPE & SQA & MMMU & SEED$^{I}$ & VizWiz & RelAcc. \\
\midrule
\multicolumn{10}{c}{\cellcolor{gray!20}\textit{Upper Bound (100\%)}} \\
\midrule
Baseline & 62.3 & 79.1 & 2318 & 87.9 & 84.7 & 51.1 & 76.5 & 68.2 & 100\% \\
\midrule
\multicolumn{10}{c}{\cellcolor{gray!20}\textit{Token Retention Ratio 22.2\% ($\downarrow$ 77.8\%)}} \\
\midrule
PACT\,{\scriptsize\textit{CVPR-25}}       & 55.8 & 72.2 & 2012 & 82.4 & 78.3 & 48.8 & 71.7 & 67.4 & 92.8\% \\
OccamToken &    61.3  &    77.1  &  2157    &    86.8  &  81.4    &  47.7     &   74.1   &     67.1 &     \textbf{96.5\%}   \\
\midrule
\multicolumn{10}{c}{\cellcolor{gray!20}\textit{Token Retention Ratio 11.1\% ($\downarrow$ 88.9\%)}} \\
\midrule
PACT\,{\scriptsize\textit{CVPR-25}}       & 50.1 & 63.1 & 1785 & 71.4 & 75.0 & 48.5 & 66.0 & 63.1 & 85.8\% \\
OccamToken &    60.2  &  76.7    &   2096   &   84.1   &   77.5   &  46.9    &   70.6   &   64.8   &     \textbf{93.8\%}    \\
\bottomrule
\end{tabular}%
}
\vspace{5pt}
\caption{Performance comparison on Qwen2-VL 7B under different token retention ratios.}

\label{tab:onevision}
\end{table}

\begin{table}[t]
\centering
\small
\setlength{\tabcolsep}{4pt}
\resizebox{0.82\textwidth}{!}{%
\begin{tabular}{l|cccccc|c}
\toprule
Method & GQA & SQA$^{I}$ & VQA$^{T}$ & POPE & MME & MMB & RelAcc. \\
\midrule
\rowcolor{gray!20} \multicolumn{8}{c}{\textit{Upper Bound, 576 Tokens (100\%)}} \\
\midrule
Vanilla & 63.2 & 72.8 & 61.3 & 85.9 & 1818 & 67.7 & 100\% \\
\midrule
\rowcolor{gray!20} \multicolumn{8}{c}{\textit{Retain 128 Tokens ($\downarrow$ 77.8\%)}} \\
\midrule
VisionZip   & 57.9 & 74.0 & 58.7 & 85.2 & 1743 & \textbf{66.7} & 97.1\% \\
OccamToken  & \textbf{61.2} & \textbf{74.1} & \textbf{60.1} & \textbf{85.5} & \textbf{1798} & 66.2 & \textbf{98.8\%} \\
\midrule
\rowcolor{gray!20} \multicolumn{8}{c}{\textit{Retain 64 Tokens ($\downarrow$ 88.9\%)}} \\
\midrule
VisionZip   & 56.2 & \textbf{74.4} & 57.4 & 76.0 & 1676 & 64.9 & 93.6\% \\
OccamToken  & \textbf{59.3} & 73.9 & \textbf{59.6} & \textbf{85.4} & \textbf{1781} & \textbf{65.8} & \textbf{97.9\%} \\
\bottomrule
\end{tabular}%
}
\caption{
Performance comparison on LLaVA-v1.5 13B under different token retention budgets.
OccamToken consistently outperforms VisionZip at the same average token budget, with especially large gains under the 64-token setting.
These results show that register-anchored adaptive pruning remains effective for larger VLM backbones.
}
\label{tab:llava15-13b}
\end{table}

\subsection{Results on Qwen-VL 7B}

\textbf{Qwen2-VL 7B.}
Table~\ref{tab:onevision} answers the architecture-transfer question by evaluating OccamToken on Qwen2-VL, whose native dynamic-resolution visual encoder does not provide an explicit \texttt{[CLS]} token.
This setting tests whether our method depends on a specific evaluator token or whether the register-anchored relative comparison is itself transferable.
We instantiate Stage~I with the mutual-scoring scheme in Section~\ref{sec:stage1}, while keeping the register token as the dynamic reference anchor.
OccamToken achieves \textbf{96.5\%} relative accuracy at 22.2\% retention and \textbf{93.8\%} at 11.1\% retention, outperforming PACT by 3.7 and 8.0 points, respectively.
The larger gap under tighter compression further supports our analysis: as the pruning boundary becomes more critical, comparing visual tokens against an internal register reference is more robust than relying on a static retention ratio.
These results show that OccamToken is not tied to \texttt{[CLS]}-based scoring and can generalize to different visual encoder designs.

\subsection{Results on LLaVA-v1.5 13B}
\textbf{LLaVA-v1.5 13B.}
Table~\ref{tab:llava15-13b} reports additional results on LLaVA-v1.5 13B under two token retention budgets.
This setting evaluates whether OccamToken remains effective when applied to a larger language backbone while using the standard 576-token visual prefix.
Compared with the 7B setting, the 13B model provides stronger language reasoning capacity, making it a useful testbed for examining whether visual token pruning can preserve the benefits of model scaling.
OccamToken consistently achieves a better accuracy--efficiency trade-off than VisionZip.
At the 128-token setting, OccamToken preserves 98.8\% relative accuracy, outperforming VisionZip by 1.7 percentage points while retaining the same average number of visual tokens.
It improves over VisionZip on most benchmarks, including GQA, SQA$^{I}$, VQA$^{T}$, POPE, and MME, showing that register-anchored pruning preserves both general visual reasoning and fine-grained visual evidence.
The advantage becomes more pronounced under stronger compression.
When retaining only 64 tokens on average, OccamToken maintains 97.9\% relative accuracy, whereas VisionZip drops to 93.6\%.
This corresponds to a 4.3-point gain in relative accuracy under the same token budget.
The improvement is particularly large on POPE and MME: OccamToken improves POPE from 76.0 to 85.4 and MME from 1676 to 1781.
These results suggest that fixed or similarity-based compression can discard visually critical evidence under tight budgets, while register-anchored relative comparison better preserves tokens that remain informative beyond the global reference.
Overall, the LLaVA-v1.5 13B results further confirm that OccamToken is not restricted to 7B-scale VLMs.
It remains effective for larger backbones and continues to provide robust performance under aggressive visual token compression.

\section{Broader Impact}
\label{sec:bimpact}
OccamToken improves the inference efficiency of VLMs by reducing redundant visual tokens without additional training. 
This can lower computational cost, memory usage, and energy consumption, making multimodal models more practical for resource-constrained deployment. 
However, efficiency improvements may also accelerate the deployment of VLMs in sensitive scenarios. 
Moreover, token pruning may affect model reliability when critical visual evidence is removed. 
Therefore, OccamToken should be carefully evaluated before being used in safety-critical applications and should inherit the same safeguards required for the underlying VLMs.

 \begin{figure}[t]
  \centering
  \includegraphics[width=1\textwidth]{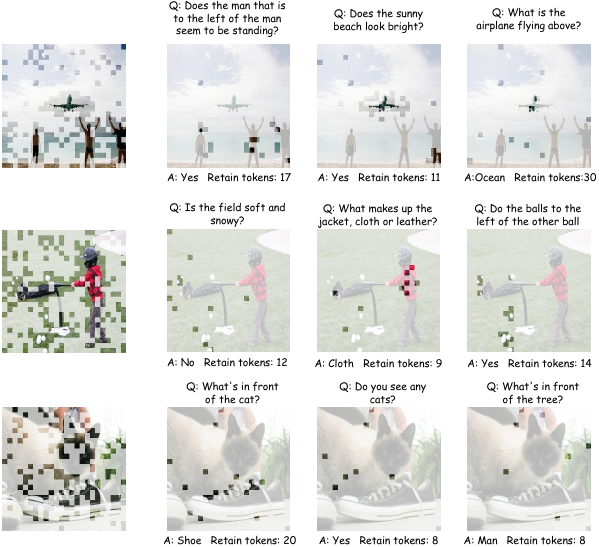}
\caption{
\textbf{Visualization of query-adaptive token budgets.}
Given the same image, OccamToken retains different visual tokens and different token counts for different queries.
This shows that the final budget is determined by query-specific evidence demand rather than a fixed top-$K$ rule.
}
  \label{fig:query_adaptive_appendix}
\end{figure}

\clearpage

 \begin{figure}[t]
  \centering
  \includegraphics[width=0.8\textwidth,height=0.74\textheight,keepaspectratio]{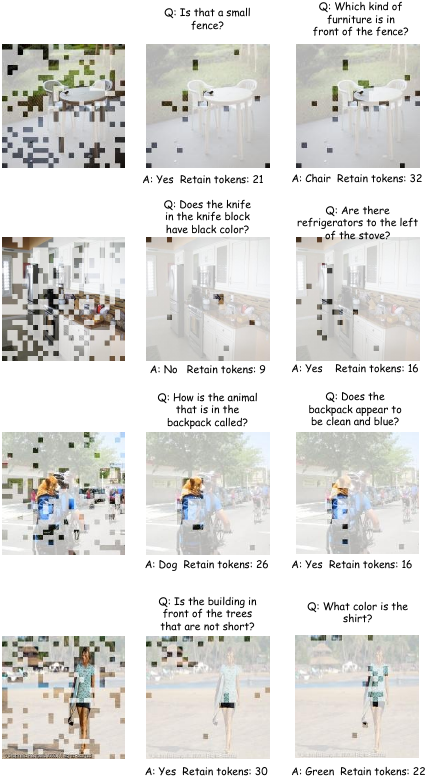}
  \caption{
\textbf{Additional visualization of query-adaptive token budgets.}
Each row shows the same image with different textual queries and the visual tokens retained by OccamToken.
Although the image content is shared, the retained regions and token counts vary across questions, reflecting different evidence demands for object existence, attributes, categories, and spatial relations.
This shows that OccamToken does not impose a fixed top-$K$ budget, but selects query-dependent visual evidence after image-level redundancy pruning.
The examples qualitatively demonstrate that Stage-II pruning adapts the final token budget to the current image--query pair.
}
  \label{fig:query_adaptive_appendix_more}
\end{figure}

\clearpage
\begin{figure}[t]
  \centering
  \includegraphics[width=1\textwidth,height=0.78\textheight,keepaspectratio]{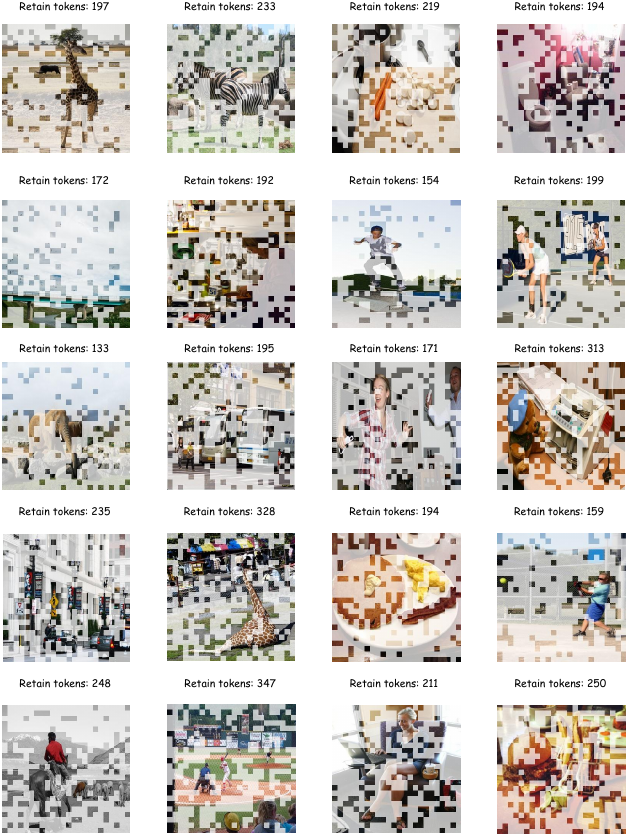}
\caption{
\textbf{Visualization of Stage-I image-adaptive redundancy pruning.}
We visualize the visual tokens retained after the first pruning stage across diverse images.
Stage-I removes image-level redundant tokens before the LLM while preserving tokens distributed over informative objects, foreground regions, and scene context.
The retained token counts vary across images, showing that OccamToken adapts the intermediate visual budget to image-specific redundancy rather than enforcing a fixed top-$K$ allocation.
}
  \label{fig:stage1_visualization}
\end{figure}

\end{document}